\documentclass[lettersize,journal]{IEEEtran}  
\IEEEoverridecommandlockouts                             

\def\BibTeX{{\rm B\kern-.05em{\sc i\kern-.025em b}\kern-.08em
		T\kern-.1667em\lower.7ex\hbox{E}\kern-.125emX}}
\usepackage{balance}
\usepackage[colorlinks,
bookmarksopen,
bookmarksnumbered,
citecolor=black,
linkcolor=black,
urlcolor=magenta]{hyperref}
\usepackage{moreverb,url}
\usepackage{amssymb}
\usepackage{mathtools}
\usepackage{mathrsfs}
\usepackage[font=small, justification=justified]{caption}
\usepackage{booktabs}
\usepackage{multirow}
\usepackage[table]{xcolor}
\usepackage[ruled,linesnumbered]{algorithm2e}
\usepackage{algpseudocode}
\usepackage{subfigure}
\usepackage{float}
\usepackage{bbm}
\usepackage{soul, xcolor}
\usepackage{xcolor}
\usepackage{diagbox}
\usepackage{threeparttable}
\usepackage{arydshln}
\usepackage{mathabx}
\usepackage{siunitx}
\usepackage{amsthm}
\usepackage{cite}

\SetCommentSty{mycommfont}

\allowdisplaybreaks

\newtheorem{remark}{Remark}

\title{\LARGE \bf
Autonomous Tail-Sitter Flights in Unknown Environments
}

\author{
Guozheng Lu$^{*1}$, Yunfan Ren$^{*1}$,  Fangcheng Zhu$^{1}$, Haotian Li$^{1}$, Ruize Xue$^{1}$,  Yixi Cai$^1$,  \\ Ximin Lyu$^{2}$ and Fu Zhang$^{1}$
\thanks{*These two authors contributed equally to this work..}
\thanks{$^{1}$The authors are with the Department of Mechanical Engineering, The University of Hong Kong, Hong Kong. {\tt\small \{gzlu, renyf,  zhufc, haotianl, u3603390, yicicai\}@connect.hku.hk, fuzhang@hku.hk}. $^{2}$ X. Lyu is with the School of Intelligent System Engineering, Sun Yat-sen University, Shenzhen, China. {\tt\small lvxm6@mail.sysu.edu.cn}.}%
}

\begin{document}

\maketitle

\begin{abstract}
Trajectory generation for fully autonomous flights of tail-sitter unmanned aerial vehicles (UAVs)  presents substantial challenges due to their highly nonlinear aerodynamics. In this paper, we introduce, to the best of our knowledge, the world's first fully autonomous tail-sitter UAV capable of high-speed navigation in unknown, cluttered environments. The UAV autonomy is enabled by cutting-edge technologies including LiDAR-based sensing, differential-flatness-based trajectory planning and control with purely onboard computation. In particular, we \textcolor{black}{propose an optimization-based} tail-sitter trajectory planning framework that generates high-speed, collision-free, and dynamically-feasible trajectories. To efficiently and  reliably solve this nonlinear, constrained \textcolor{black}{problem}, we develop an efficient feasibility-\textcolor{black}{assured} solver, EFOPT, tailored for \textcolor{black}{the online planning of tail-sitter UAVs}. We conduct extensive simulation studies to benchmark EFOPT's superiority in planning tasks against conventional NLP solvers.  We also demonstrate exhaustive experiments of aggressive autonomous flights with speeds up to \SI{15}{m/s} in various real-world environments, including indoor  laboratories, underground parking lots, and outdoor parks. A video demonstration is available at \url{https://youtu.be/OvqhlB2h3k8}, and the EFOPT solver is  open-sourced at \url{https://github.com/hku-mars/EFOPT}.
\end{abstract}

\section{Introduction}
In recent years, autonomous unmanned aerial vehicles (UAVs) in conventional fixed-wing and rotary-wing configurations have been pervasively used  in  a broad range of applications, such as  geographical surveys,  agriculture monitoring and spraying, and tunnel inspections. However, these two platforms have inherent limitations that restrict their practical use. Fixed-wing airplanes excel in aerodynamic efficiency and fast flight in open space, but \textcolor{black}{lack flexibility of hovering and slow-speed flights,} and typically rely on runways for takeoff and landing.  Conversely, rotary-wing vehicles like multicopters, \textcolor{black}{present} remarkable agility in hovering, vertical takeoff and landing (VTOL),  and aggressive maneuvers, \textcolor{black}{allowing applications in complex environments}. However,  they are generally less energy-efficient for long-range missions.

The hybrid  design, \textcolor{black}{combining} fixed and rotary-wing elements,  offers a complementary solution that boasts both  the VTOL capability and aerodynamic efficiency in high-speed flights. These hybrid VTOL UAVs are ideally suited for tasks in diverse environments.  The potential interest has spurred a variety of VTOL UAV designs, including tilt-rotors \cite{carlson2014hybrid, ozdemir2014design}, tilt-wings \cite{ccetinsoy2011design}, rotor-wings \cite{mckenna2007one}, dual-systems \cite{park2014arcturus, gu2017development} and tail-sitters \cite{de2018design, gu2018coordinate, tal2021global, ritz2017global}. Among these, tail-sitters have rotors fixed to the wings and utilize their thrust throughout the entire flight regime,  eliminating the need for extra tilting mechanisms or redundant propulsion.  This feature of mechanical simplicity is especially crucial for small-scale, lightweight, low-cost, portable, and efficient UAVs. 

\begin{figure}
	\centering
	\includegraphics[width=1\linewidth]{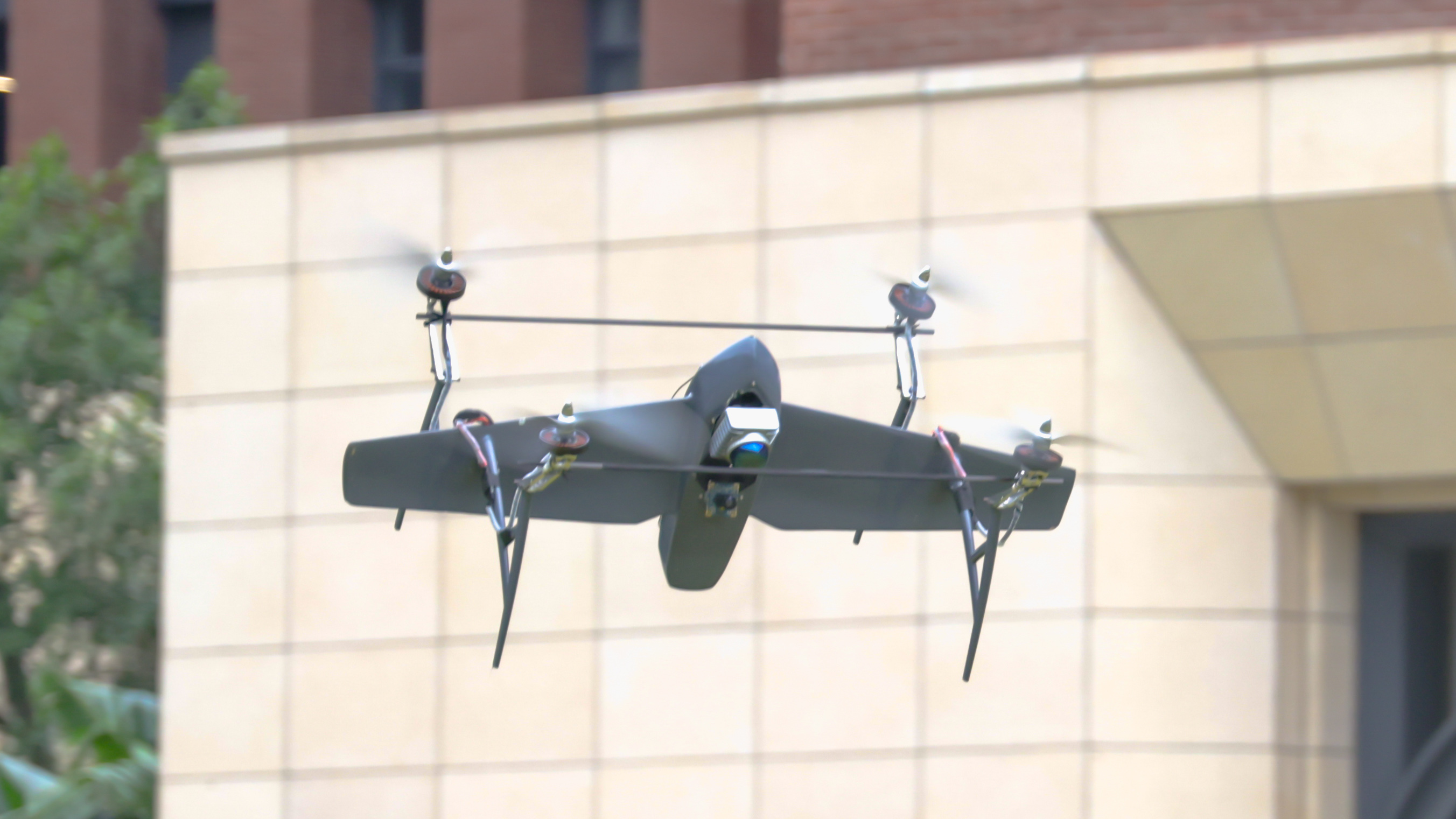} 
	\caption{Autonomous tail-sitter UAV equipped with a LiDAR, driven by real-time onboard perception, planning, and control.} 
	\label{fig_perface}
\end{figure}

Despite the concept of tail-sitters has been around for decades, they are rarely seen in practical use, primarily due to control difficulties. Typically, a tail-sitter takes off and lands vertically on its tail, and transitions to a nearly horizontal attitude for forward flight. This transition induces significant variations in aerodynamic forces and moments acting on the vehicle. These aerodynamics are highly nonlinear functions of attitude and airflow, posing substantial challenges for flight controllers to perform highly maneuverable flights. \textcolor{black}{Recent progress in tail-sitter controls has exploited aerodynamic models within real-time advanced controllers \cite{ritz2017global, smeur2020incremental, tal2022global, lu2024trajectory}, enabling successful demonstrations of aggressive flights. However,  these controllers track reference trajectories that are either low-dimensional course lines or finely-tuned trajectories designed offline. Compared to conventional multicopter and fixed-wing UAVs, tail-sitter planning is significantly more challenging due to their extremely nonlinear aerodynamics. A common limitation of existing studies on tail-sitter planning is that their algorithms rely on extensive offline computation.}

\textcolor{black}{The agile maneuverability of tail-sitter UAVs and the precise flight control required for autonomous navigation necessitate high-dimensional trajectories that are smooth, fast, and responsive to \textcolor{black}{complex} environments, while respecting flight dynamics and actuator constraints. Tail-sitter maneuvers are mostly generated through trajectory optimization, which can approach task-specified goals while considering constraints imposed by obstacle avoidance and dynamic feasibility.  However,  incorporating high-fidelity aerodynamic models into the optimization to ensure high control performance, typically leads to nonlinear programming (NLP) problems that are non-convex and generally difficult to solve. The computational complexity conflicts with the demand for high-speed tail-sitter navigation, which requires feasible trajectories computed in short time intervals to make response to \textcolor{black}{online-perceived} environments.  As online planning for tail-sitters remains an open question, no tail-sitter UAV capable of fully autonomous navigation in real-world environments has yet been demonstrated.}

\subsection{Contributions}
In this paper, we present a fully autonomous tail-sitter UAV, as shown in Fig. \ref{fig_perface}, which can navigate through unknown, obstacle-dense environments at high speeds.  The key contributions of this paper are as follows:
\begin{itemize}
	\item [1.]  We present an optimization-based trajectory planning framework  that generates high-speed, collision-free, and dynamically-feasible trajectories  for tail-sitters.
	
	\item [2.] To address real-time trajectory optimization, we develop an  \textbf{E}fficient \textbf{F}easibility-\textcolor{black}{assured} \textbf{OPT}imization solver,\textbf{ EFOPT}, \textcolor{black}{based on the $\ell_1$  penalty method and sequential quadratic programming (SQP). We  benchmark the proposed EFOPT against state-of-the-art  NLP solvers on  extensive tail-sitter trajectory optimization problems.}
	
	\item [3.]  \textcolor{black}{We develop an autonomous tail-sitter UAV prototype and detail the complete system pipeline of onboard perception, planning, and control. We demonstrate, to the best of our knowledge, the world's first fully autonomous tail-sitter navigation in  unknown environments at high speeds.}
\end{itemize}

\subsection{Outline}
The outline of the rest of the paper is as follows. Section \ref{sec_relatedWork} reviews related works. \textcolor{black}{Section \ref{sec_traj_opt}  presents the trajectory optimization formulation for tail-sitter UAVs. A novel NLP solver tailored for online trajectory optimization is proposed in Section \ref{sec_EFOPT}. Section V overviews the algorithm pipeline of the autonomous system.} Section \ref{sec_benchmark} benchmarks the proposed solver with off-the-shelf NLP solvers in the context of trajectory optimization, followed by real-world experiments in Section \ref{sec_experiment}. Limitations and potential extensions of the proposed framework are discussed in Section \ref{sec_discuss}. Finally, \textcolor{black}{the} conclusion of the paper arrives at Section \ref{sec_conclusion}.

\section {Related Works}
\label{sec_relatedWork}
\subsection{\textcolor{black}{Tail-sitter trajectory generation}}
Trajectory generation methods for tail-sitters  can generally be categorized into two types based on two main control strategies: the separated control strategy, which involves  distinct trajectory generation and  controllers for different flight phases, and the global control strategy,  which regulates the vehicle maneuvers throughout the entire flight envelope under a unified framework.

The separated control strategy \cite{frank2007hover, oosedo2013development, lyu2017hierarchical} typically divides the tail-sitter flights into three phases: vertical flight (e.g., takeoff, landing, and hovering), transition, and level flight. During \textcolor{black}{the} low-speed vertical flight, vehicle dynamics reduce to rotary-wing models, while fixed-wing models are applied in level flight, allowing the use of well-established trajectory generation methods for multicopters \cite{mellinger2011minimum} and fixed-wing UAVs \cite{park2004new}. However, designing transition maneuvers between hovering and level flight is challenging due to complex aerodynamics. During the transition, significant changes in angle of attack (AoA) and airspeed make the aerodynamics highly nonlinear. A straightforward linear transition method is feeding a pre-designed profile of linearly varying pitch angle to the attitude controller with a constant altitude command \cite{verling2016full, lyu2017design}. This simple combination of commands is often dynamically infeasible, causing substantial height deviations that require additional sophisticated controllers for error correction \cite{xu2019acceleration}. To enhance control accuracy, high-fidelity aerodynamic models and actuator saturation are incorporated into trajectory optimization frameworks. For example, Naldi and Marconi \cite{naldi2011optimal} addressed a time-and-energy-optimal transition problem, while Oosedo et al. \cite{oosedo2017optimal} and Li et al. \cite{li2020transition} further included altitude constraints during transitions. However, these  trajectory optimizations are non-convex and limited to offline implementation. Additionally, studies on transition planning only consider 2-D longitudinal motion, preventing transition with lateral banked turns for obstacle avoidance. McIntosh et al. \cite{mcintosh2022transition} considered longitudinal evasive maneuvers during transition, but the application scenarios are limited. Overall, existing literature on the separated strategy either omits dynamic feasibility by oversimplified models or restricts the flight envelope to simplify planning. Consequently, these solutions and their computational demands hinder real-time planning in real, cluttered environments.

Despite the simple implementation of  the separate control strategy for tail-sitter prototype validation, especially using commercial autopilots \cite{meier2015px4}, its limitations--such as controller switching that introduces undesired transient response and  confined maneuverability in trajectory generation--restrict its applicability. In contrast, the global  strategy, while utilizing a unified dynamic model for both trajectory generation and control without mode switching, shows greater promise for practical applications.

Research on global trajectory planning for tail-sitters spans various levels of model fidelity, but none have practically demonstrated online planning in cluttered environments. \textcolor{black}{Smeur et al. \cite{smeur2020incremental} used a proportional-derivative (PD) controller to generate acceleration reference to connect} \textcolor{black}{waypoints and course lines, leaving aerodynamic compensation to an incremental nonlinear dynamic inversion (INDI) controller, which uses inertial measurement unit (IMU) data and feedback linearization to estimate and compensate for aerodynamics and other uncertainties. However, INDI adaptation is limited by sensor noise, vibrations, filter delay, model fidelity, and external winds, resulting in notable altitude errors of around \SI{2}{m} in forward flight \cite{smeur2020incremental}. Moreover,  PD gains are kept conservative to maintain the convergent margin of the INDI linearization, restricting this approach to low-agility maneuvers.} McIntosh et al. \cite{mcintosh2021optimal} considered  longitudinal dynamics using a first-order aerodynamic model and proposed a trajectory optimization that is non-convex and solved offline. Similarly, Tal et al. \cite{tal2023aerobatic}  derived a simplified differential flatness based on the  $\phi$-theory model \cite{lustosa2017phi}, and developed a trajectory optimization that sequentially minimizes time allocation and snap energy. By employing INDI techniques to compensate for modest model inaccuracies, the approach generates dynamically feasible trajectories for uncoordinated flights and agile aerobatics but relies on computationally intensive solvers unsuitable for onboard implementation. Our recent work \cite{lu2024trajectory} theoretically proved the tail-sitter differential flatness using classic aerodynamic models without simplification, and  demonstrated high-accuracy, real-time control for aggressive flights. Nevertheless, our trajectory generation, which incorporates time allocation and dynamic feasibility in a single optimization, results in a non-convex constrained problem. Although the computational efficiency has been improved to hundreds of milliseconds on Intel i7 processors, it remains insufficient for online planning. To date, no existing scheme is capable of real-time trajectory planning for aggressive tail-sitter flights.

\subsection{Autonomous fixed-wing and multicopter UAVs}
Compared to tail-sitters, planning and control techniques for fixed-wing and multicopter UAVs are well-developed due to their relatively simpler aerodynamics. In consequence, a wealth of autonomy applications have been developed for these UAVs, which are potentially adaptable to tail-sitters.

Fixed-wing airplanes typically operate within a conservative level flight regime, where the wing aerodynamic forces are approximately linear. Fixed-wing guidance often simply considers flight path geometry by connecting line and arc segments (e.g. L1 guidance \cite{park2004new}) and assigns aircraft heading tangent to the curve. The trajectory feasibility is simply approximated as curvature constraints for cruise flight in open space at high altitudes. To achieve aggressive fixed-wing flights in complex  environments, Bry et al. \cite{bry2015aggressive} and Majumdar and Tedrake \cite{majumdar2017funnel} considered obstacle avoidance and control feasibility in trajectory optimizations, which, however, rely on offline computation.



Multicopters have simpler dynamics and distinctive maneuverability, allowing more efficient trajectory planning onboard to achieve autonomous flights with minimal human intervention. Leveraging the differential flatness property  \cite{mellinger2011minimum, faessler2017differential}, trajectory generation for multicopters has been extensively studied  and achieved a broad class of applications, such as flying through narrow gaps \cite{mellinger2011minimum, falanga2017aggressive}, perching on structures \cite{mellinger2011minimum, hang2019perching, hsiao2023energy}, catching a thrown ball \cite{mueller2015computationally} and drone racing \cite{kaufmann2023champion}. Flying without a priori maps and external motion capture systems, autonomous UAVs also necessitate online perception using exteroceptive sensors. Multicopters equipped with cameras have achieved fully autonomous navigation and exploration \cite{liu2016high, zhou2021fuel} in the real world, by integrating techniques of visual-inertial odometry \cite{forster2016manifold, qin2018vins}, simultaneous localization and mapping (SLAM) \cite{campos2021orb}, and online planning algorithms \cite{zhou2020ego, zhou2019robust, loquercio2021learning}. However, visual perception typically degrades in fast motion (e.g., speed over \SI{10}{m/s}) due to the camera's inherent limitations of motion blur and changing illumination. In contrast, lightweight LiDAR sensors provide dense, accurate and low-latency 3-D point clouds that are more robust for high-speed flights. Recent research on real-time LiDAR-based odometry and SLAM \cite{xu2022fast} has enabled multicopters to navigate much faster in complex environments \cite{ren2022bubble, wang2022geometrically, ren2022online}.

In terms of online trajectory planning for multicopters, existing works can be categorized into four main approaches: search-based, sampling-based, optimization-based and learning-based methods. Search-based methods, such as Dijkstra's \cite{dijkstra1959note} and A* \cite{hart1968formal} algorithms, which search over a graph representing feasible action primitives, have shown promising results in autonomous flights \cite{liu2018search, zhou2019robust, zhang2020falco}. However, the computational expense of graph search escalates with the increase of state dimension,  due to the exponential growth in node numbers. Thus, existing online planning typically considers kinematic primitives rather than true actuator actions.  Sampling-based planning algorithms like  Probabilistic Roadmap (PRM) \cite{kavraki1996probabilistic},  Rapidly- exploring Random Tree (RRT) \cite{lavalle1998rapidly} and their variants \cite{bohlin2000path}, \cite{karaman2011sampling},  spread sparse random samples on the graph and connect  selected ones to seek collision-free paths in configuration space, showing exceptional scalability for large maps.  However, in cluttered environments,  the computational efficiency of these methods is not guaranteed due to the dependence on map complexity. Optimization-based methods focus on planning trajectory by solving optimal control problems. Instead of directly optimizing the control sequence using shooting \cite{houska2011acado} or collocation methods \cite{patterson2014gpops},  real-time planners usually leverage differential flatness properties and efficient trajectory parameterizations, such as polynomials \cite{mellinger2011minimum}, B-splines \cite{tang2021real} and  B\'ezier curves \cite{gao2020teach}, to enhance computational efficiency. Obstacle information is typically encoded as constrained terms in the optimization, such as safe corridors \cite{gao2020teach, ren2022bubble} and  Euclidean signed distance field  \textcolor{black}{(ESDF) \cite{usenko2017real}.  While ESDF usually results in non-convex constraints due to its direct representation of distance to the nearest obstacle, safe-corridor methods that construct obstacle-free space as convex hull \cite{deits2015computing, zhong2020generating, toumieh2022voxel}  can effectively reduce the optimization complexity.} Recently, machine learning has also demonstrated exceptional results in training end-to-end control policies directly from flight data \cite{loquercio2021learning}, but it usually requires high-fidelity simulators and techniques for simulation-to-real transfer.

\subsection{Trajectory optimization methods}
In this paper, we concentrate on optimization-based trajectory planning methods. \textcolor{black}{Unlike aggressive tail-sitter maneuvers,  most online trajectory optimization approaches for multicopters only consider kinematic constraints at moderate speeds, omitting vehicle dynamics for computational efficiency.} With heuristic time allocation, trajectories can be efficiently computed by convex optimization, such as quadratic programming (QP) \cite{mellinger2011minimum},  mixed-integer quadratic programming (MIQP) \cite{tordesillas2019faster}, and second-order cone programming (SOCP) \cite{gao2020teach}.  To address infeasible solutions caused by improper time allocation, temporal variables can be incorporated into trajectory optimizations but result in non-convex problems. \textcolor{black}{Alternative optimization procedures that iteratively solve sub-problems dependent on separated spatial and temporal variables, have been proposed \cite{richter2016polynomial, gao2020teach, sun2021fast} for real-time computation. However, this scheme lacks theoretical convergence guarantees and robustness for complex problems.} Wang et al. \cite{wang2022geometrically} proposed a novel polynomial parameterization that allows simultaneous spatial-temporal optimization in real-time, but constraints are relaxed as soft penalties to reduce the NLP complexity. \textcolor{black}{Although these optimization methods have been successfully applied to online planning for multicopters, they are less effective for tail-sitters. As simple kinematic models and heuristic time allocation cannot ensure the dynamic feasibility required for fast maneuvering tail-sitters, trajectory optimizations should incorporate both flight dynamics and time allocation. Tal et al. \cite{tal2023aerobatic} and our previous work \cite{lu2024trajectory}  respectively adopted alternative optimization and soft penalty methods  to reduce computational complexity in trajectory optimization for aggressive tail-sitter flights. However, their demonstrations rely on cumbersome tuning for different scenarios and excessive computational load, showing these methods are unreliable for online planning.}

Offline trajectory optimization for tail-sitters \cite{oosedo2017optimal, li2020transition, mcintosh2021optimal} usually takes the problem as a general NLP and solve by standard solvers, such as SNOPT \cite{snopt} and IPOPT \cite{ipopt}. \textcolor{black}{These general-purpose solvers enclosure mathematical algorithms and heuristics that improve the overall success, but may struggle with specific complex problems. In contrast, customized optimizers, tailored to task-specific information and requirements, often excel in certain cases. Hierarchical quadratic programming (HQP) \cite{escande2014hierarchical} ensures strict satisfaction with higher-priority constraints while relaxing lower-priority ones, to achieve real-time task optimization for  humanoid and quadrupedal controls \cite{hutter2014quadrupedal, feng2015optimization, feng2016online}. However, HQP is limited to standard QPs.} Besides, Schulman et al.\cite{schulman2014motion} presented TRAJOPT, \textcolor{black}{an} $\ell_1$ penalty method with sequential convex programming (SCP) to ensure collision-free constraint for manipulator motion planning. Although the convex approximation and  adaptive constraint penalty are well-suited to handle the nonlinearity and hard constraints, TRAJOPT is less effective in tail-sitter planning, as demonstrated in the benchmark study in Section \ref{sec_benchmark}.

\section{Tail-Sitter Trajectory Optimization Formulation}
\label{sec_traj_opt}
\begin{figure}[t!] 
    \centering
    \includegraphics[width=1\linewidth]{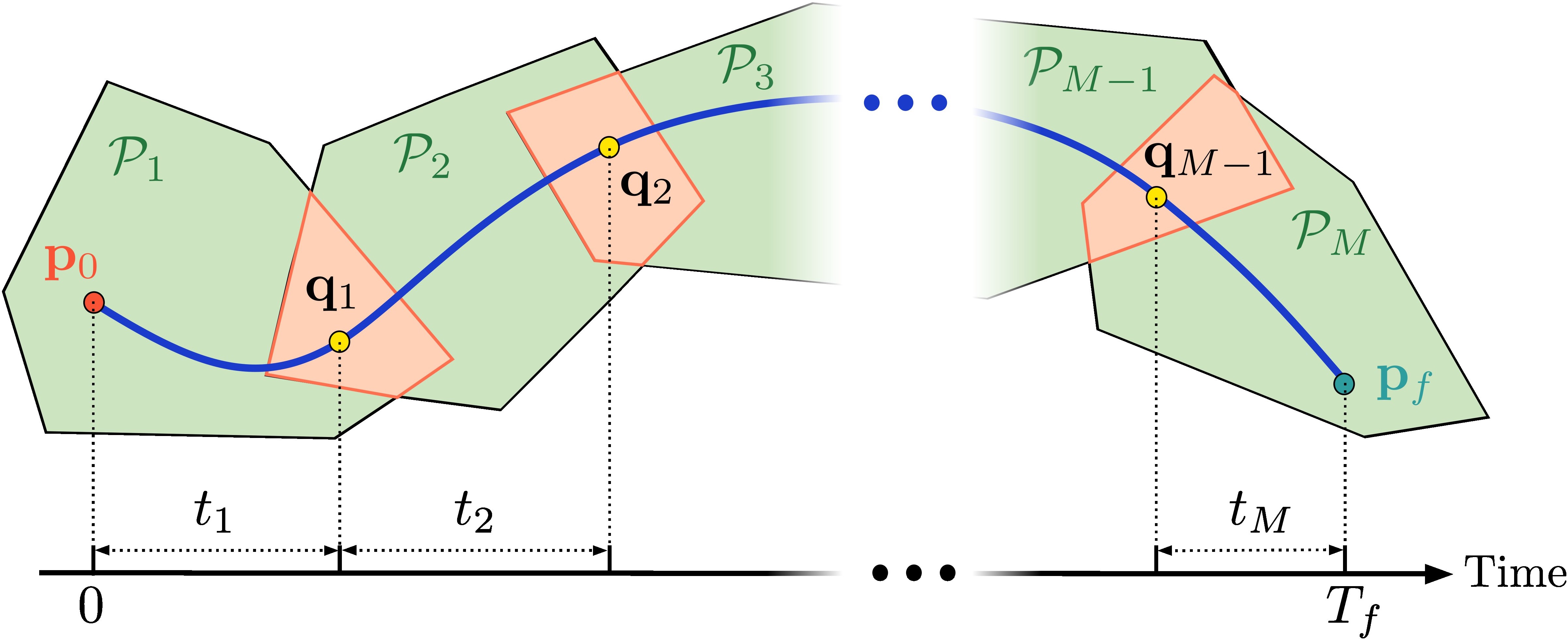} 
    \caption{Trajectory paramterization. The trajectory (blue curve) consecutively connects the initial position $\mathbf p_0$, internal waypoint sequence $\mathbf q_1, \cdots, \mathbf q_{M-1}$, and terminal position $\mathbf p_f$. Each trajectory segment has flight duration $t_i$, and the entire flight duration is $T_f$.} 
    \label{fig_traj_param}
\end{figure}

\textcolor{black}{In our previous work \cite{lu2024trajectory}, we found tail-sitter is a  differentially flat system throughout the entire flight envelope, by specifying the coordinate flight condition and choosing the vehicle position as the flat output. All the system states and inputs can be expressed by functions of position and its derivatives, based on accurate aerodynamic models without simplifications. This property greatly reduces the trajectory planning problem to low-dimensional algebra in the flat-output space (i.e., Cartesian space). The tail-sitter flight dynamics and differential flatness are briefly introduced in Appendix \ref{app_dynamics} and \ref{app_diff_flat}.}

Based on the tail-sitter differential flatness, we formulate the trajectory planning problem as an optimization to generate a trajectory $\mathbf{p}(t): \mathbb{R} \in [0, T_f] \mapsto \mathbb{R}^3$ that is dynamically feasible and collision-free, while minimizing the flight time $T_f$ and snap energy $\|\mathbf{p}^{(4)}\|^2$ \textcolor{black}{for smoothness}.  The dynamic feasibility and obstacle avoidance are guaranteed by respectively incorporating the actuator saturation constraints and positional constraints of safe flight corridors into the optimization.

We parameterize the trajectory as piecewise polynomials using a spatial-temporal deformation proposed in \cite{wang2022geometrically}. Given a flight corridor $\mathcal P$ comprised of $M$ consecutively overlapped convex polyhedra $\mathcal P_i$, both the trajectory $\mathbf p(t)$ and flight duration $T_f$ are divided into $M$ segments, as shown in Fig. \ref{fig_traj_param}. Denote $\mathbf p(t) = (\mathbf p_1(t), \cdots, \mathbf p_M(t))$, and  $\mathbf T = (t_1, \cdots, t_M)$ where $\sum_{i = 1}^{M} t_i = T_f$. We introduce \textcolor{black}{a} sequence of internal waypoints $\mathbf Q = (\mathbf q_1, ..., \mathbf q_{M-1})$. Each waypoint element is inserted at each intersection of the polyhedra, ensuring $\mathbf q_i  \in (\mathcal P_i \cap \mathcal P_{i+1})$. Each segment of the trajectory $\mathbf p_i (t)$ is confined within the $i$-th polyhedron $\mathcal P_i$, connecting $\mathbf q_i$ (or $\mathbf p_0$) and $\mathbf q_{i+1}$ (or $\mathbf p_f$). These trajectory segments create a continuous path from the initial position $\mathbf p_0$, through the waypoints sequence $\mathbf Q$, to the terminal position $\mathbf p_f$. As suggested by \cite{wang2022geometrically}, this piecewise trajectory is expressed as a set of minimum-snap polynomials with $C^3$ continuity. Given the initial state $\mathbf{s}_0$, terminal state $\mathbf{s}_f$, the piecewise polynomial trajectory is eventually parameterized by:
\begin{equation}
	\mathbf p_i(t) = \mathbf{C}_i\left(\mathbf Q, \mathbf T, \mathbf s_0, \mathbf s_f \right) \beta(t), \quad t \in [0, t_i]
	\label{e_poly}
\end{equation}
where $\mathbf C_i$ is a polynomial coefficient matrix related to internal waypoint $\mathbf Q$, time allocation $\mathbf T$, and given initial and terminal states \textcolor{black}{$\mathbf{s}_0$}, $\mathbf{s}_f$. $\beta(t) = [t^0, \cdots, t^7]^T$.

The trajectory optimization based on this polynomial parameterization is then formulated as follows: 
\begin{subequations}
    \label{e_traj_opt}
	\begin{align}
		&\displaystyle\min_{\mathbf Q, \mathbf T} \displaystyle \int_{0}^{T_f}  \Vert \mathbf{p}^{(4) } (t) \Vert^2 dt + \rho T_f  \label{e_opt_cost}\\
		&\mathrm{s.t.} \ \mathbf p(t) = \mathbf{C}(\mathbf Q, \mathbf T,  \mathbf s_0, \mathbf s_f) \beta(t), \quad T_f = \sum_{i = 1}^{M} t_i 
            \label{e_opt_poly}\\
            & \hspace{5.7mm} \mathbf{x}(t) = \mathcal{X}\!\left(\mathbf{p}^{(0:2)}(t) \right),\ \mathbf{u}(t) = \mathcal{U}\!\left(\mathbf{p}^{(1:3)}(t) \right) \label{e_opt_sys}\\
		& \hspace{5.7mm}  \mathbf p_i(t) \in \mathcal P_i, \quad i = 1, \cdots, M      \label{e_opt_sfc}\\
		& \hspace{5.7mm}  \mathbf{x}(t) \in \mathbb{X},\ \mathbf{u}(t) \in \mathbb{U} \label{e_opt_bound} \\
		& \hspace{5.7mm}  \mathcal{S}(\mathbf{x}(t)) \geq \epsilon  \label{e_opt_sing}
	\end{align}
\end{subequations}
where $\rho > 0$ is the weight of total flight time $T_f$. \textcolor{black}{$\mathcal{X} $ and $\mathcal{U}$ in (\ref{e_opt_sys}) are flatness functions given in Appendix \ref{app_diff_flat} that transform position and its derivatives into states and inputs of the UAV. $\mathcal{S}(\mathbf{x}(t)) > 0$ in (\ref{e_opt_sing}) denotes the condition for avoiding singularities (\ref{e_sing_cond}) of the flatness functions, and we implement $\epsilon = 0.1$ for numerical stability.} The constraint in (\ref{e_opt_sfc}) confines the position trajectory within the SFCs for collision safety. The constraints in (\ref{e_opt_bound}) describe the kinodynamic and actuator constraints, respectively, where:
\begin{equation}
	\Vert \mathbf{v}(t) \Vert \leq v_{\rm max} \ , \quad \Vert \dot{\mathbf{v}} (t) \Vert \leq a_{\rm max}
\end{equation}
with $v_{\rm max}$ and $a_{\rm max}$ the maximum velocity and acceleration, and $\mathbb{U} = \{\mathbf{u} \in \mathbb{R}^4 | \ \mathbf{u}_{\rm min} \leq \mathbf{u} \leq \mathbf{u}_{\rm max}\}$ is the boundary of the system inputs (i.e., the thrust acceleration $a_T$ and angular velocity $\boldsymbol{\omega}$).

The optimization problem in (\ref{e_traj_opt}) aims to optimize both the internal waypoint $\mathbf Q$ and time allocation $\mathbf T$, to pursue a time-optimal trajectory with a minimum-snap energy for smoothness, as described in the objective function (\ref{e_opt_cost}). In this paper, we assign a large weight to the flight time (i.e., $\rho = 1\text{e}4$) to pursue fast flight. However, while this high weighting enables aggressive trajectories with shorter flight duration and faster speeds, it also significantly increases  the tendency towards constraint violations, posing a substantial challenge to optimization solvers.

\section{Efficient Feasibility-\textcolor{black}{Assured} Optimization}
\label{sec_EFOPT}
The trajectory optimization in (\ref{e_traj_opt}) presents a constrained non-convex optimization problem that must be solved both efficiently and robustly to enable real-time autonomous flights. In our previous work \cite{lu2024trajectory}, we addressed offline trajectory optimization for flying through fixed waypoints without corridor constraint, using a soft penalty method combined with quasi-Newton method (e.g., LBFGS). However, as previously discussed, soft penalty methods typically require significant effort for fine-tuning penalty weights in different scenarios. More importantly, even with meticulous tuning, they often fail \textcolor{black}{by generating} solutions that are either too time-consuming or infeasible, rendering them unsuitable for online planning. 

\textcolor{black}{In the practice of online planning for autonomous flights, it is crucial to ensure trajectory feasibility where the planned trajectories have no collision with obstacles and are executable by actuators or low-level controllers.  The computational efficiency of planning is also essential to make timely response to new obstacles perceived online, especially during high-speed flights.  The trajectory quality such as short arrival time, fast flight speed and low energy consumption, is preferable but relatively less important to the task accomplishment.  Since trajectory feasibility and quality are respectively formulated as constraints and objectives, in trajectory optimization (\ref{e_traj_opt}), we propose a  tailored  NLP solver, EFOPT, that enhances computational efficiency and ensures feasibility by relaxing the optimality to some extent. The optimization is built on the $\ell_1$ penalty method and sequential quadratic programming (SQP).}


\subsection{Background: $\ell_1$ penalty method with sequential quadratic programming}
The trajectory optimization such as (\ref{e_traj_opt}) can be denoted as a general non-convex optimization:
\begin{subequations}
	\begin{align}
		&\displaystyle\min_{\mathbf x} f(\mathbf x)  \label{e_nlp_cost}\\
		&\mathrm{s.t.} \quad g_i(\mathbf x) \leq 0, \quad i = 1,2, \cdots, m \\
		&\qquad \ h_i(\mathbf x) = 0, \quad i = 1,2,\cdots, n
	\end{align}
	\label{e_nlp}
\end{subequations}
where $f, g_i, h_i$ are non-convex scalar functions. 

Sequential convex optimization is an iterative method that solves a non-convex optimization problem by repeatedly constructing a quadratic programming sub-problem, which is an approximation to the original problem around the current iterate $\mathbf x$. The original non-convex programming can be solved by iteratively calculating the QP sub-problems as follows:
 \begin{subequations}
 	\begin{align}
 		&\displaystyle\min_{\mathbf d} \tilde{f}(\mathbf x + \mathbf d)  \label{e_sqp_cost}\\
 		&\mathrm{s.t.} \quad \tilde g_i(\mathbf x + \mathbf d) \leq 0, \quad i = 1,2, \cdots, m \\
 		&\qquad \ \tilde h_i(\mathbf x + \mathbf d) = 0, \quad i = 1,2,\cdots, n
 	\end{align}
 	\label{e_sqp}
 \end{subequations}
where $\mathbf d$ is the update step and the variable vector is iteratively updated $\mathbf x \leftarrow \mathbf x + \mathbf d$ by solving QPs in (\ref{e_sqp}) until the original problem satisfies \textcolor{black}{the convergence} conditions. $\tilde{f},  \tilde{g}, \tilde{h}$ are convexified quadratic models
\begin{subequations}
    \label{e_quadratic_model}
    \begin{align}
        \tilde{f}(\mathbf x + \mathbf d) = f(\mathbf x) + \nabla f(\mathbf x)^T \mathbf d + \frac{1}{2} \mathbf d^T \left(\nabla^2 f(\mathbf x)\right) \mathbf d \\
        \tilde{g}(\mathbf x + \mathbf d) = g(\mathbf x) + \nabla g(\mathbf x)^T \mathbf d + \frac{1}{2} \mathbf d^T \left(\nabla^2 g(\mathbf x)\right) \mathbf d \\
        \tilde{h}(\mathbf x + \mathbf d) = h(\mathbf x) + \nabla h(\mathbf x)^T \mathbf d + \frac{1}{2} \mathbf d^T \left(\nabla^2 h(\mathbf x)\right) \mathbf d
    \end{align}
\end{subequations}
However, the constructed QP sub-problem in (\ref{e_sqp}) may have no feasible solutions, in the case where the quadratic approximation has considerable deviation from the original problem due to excessive non-convexity. The $\ell_1$ penalty method with SQP can be more robust to highly non-convex problems. The original problem in (\ref{e_sqp}) can be solved by iteratively solving an unconstrained minimization:
\begin{equation}
		\displaystyle\min_{\mathbf x}  f(\mathbf x) + \mu \sum_{i=1}^m |  g_i (\mathbf x) |^+ + \mu \sum_{i=1}^n |  h_i (\mathbf x) |  
	\label{e_merit_sqp}
\end{equation}
where $|a|^+ = \max(0,a)$ is $\ell_1$ penalty, and $\mu$ is a coefficient increased sequentially if (\ref{e_merit_sqp}) is solved but constraints are not satisfied. In each penalty iteration, the problem (\ref{e_merit_sqp}) is further solved by iteratively approximating the unconstrained problem by a QP with trust region constraint:
\begin{subequations}
    \label{e_trs}
	\begin{align}
		&\displaystyle\min_{\mathbf d} \tilde{f}(\mathbf x \!+\! \mathbf d) \!+\! \mu \!\sum_{i=1}^m | \tilde g_i (\mathbf x \!+\! \mathbf d) |^+ + \mu \sum_{i=1}^n |\tilde  h_i (\mathbf x \!+\! \mathbf d) |  \\
		&\mathrm{s.t.} \quad \| \mathbf{d} \| \leq s
		\end{align}		
\end{subequations}
where  $s$ is a trust region (TR) to restrict each step size such that the approximation is close enough to the original problem. Each QP in (\ref{e_trs}) is called trust-region sub-problem (TRS). 

\textcolor{black}{The penalty coefficient $\mu$ is initialized to a small value and gradually increased over iterations. Initially, the small coefficient reduces the dominance of the weighted constraint violation term in the merit function (\ref{e_trs}), enabling the search for feasible solutions to begin with lower objective costs. This property often facilitates convergence to feasible solutions with improved optimality. In the context of trajectory optimization, this approach yields trajectories with reduced time duration and/or control energy, both of which are crucial for high-speed maneuvers. In later iterations, the penalty coefficient is increased to progressively enforce constraint satisfaction.} 

\subsection{Implementation of EFOPT}
Our implementation is a variant of $\ell_1$ penalty method for sequential convex programming, presented in \cite{schulman2014motion}. The constraint violation is denoted
\begin{equation}
	C = \sum_{i=1}^m |  g_i (\mathbf x) |^+ + \sum_{i=1}^n |  h_i (\mathbf x) |
	\end{equation}
Denote the merit function as the sum of objective and weighted constraint violation as follows:
\begin{equation}
	 F(\mathbf x) = {f}(\mathbf x) + \mu C
  \label{e_merit_func}
\end{equation}
 and $\tilde F(\mathbf x)$ denotes a  convexification of the merit function in the form of quadratic models in (\ref{e_quadratic_model}). 

\begin{algorithm}[b!]
	\caption{Implementation of EFOPT }
	\small
	\label{alg_EFOPT}
	\SetAlgoLined
	\textbf{Parameters: }  \\
        \quad $m$: maximum penalty iteration number \\
        \quad $n$: maximum convexification iteration number\\
	\quad $\mu_0$: initial penalty coefficient \\  
	\quad $s_0$: initial trust region size \\
	\quad $\lambda_1, \lambda_2$: model quality thresholds ($0 < \lambda_1 < \lambda_2$) \\
	\quad $\tau^-, \tau^+$: trust region expansion and shrinkage factors ($0 < \tau^- < 1 < \tau^+$)\\
	\quad $k$: penalty scaling factor ($k > 1$) \\
	\quad xtol$^\dagger$, ftol$^\dagger$, ctol$^\dagger$: coarse thresholds \\
	\quad xtol$^{\dagger\dagger}$, ftol$^{\dagger\dagger}$, ctol$^{\dagger\dagger}$: fine thresholds $\left((\cdot)^{\dagger\dagger} < (\cdot)^\dagger \right)$  \\
	\textbf{Variables: } \\
	\quad $\mathbf x$: current solution vector \\
	\quad $\mathbf d$: update step \\
	\quad $\mu$: penalty coefficient \\
	\quad $s$: trust region size \\
        \quad xtol, ftol: convergence thresholds for $\mathbf x$ and $F(\mathbf x)$ \\
        \quad ctol: constraint tolerance  \\
        \textbf{Initialization: } \\
        \quad $\mu \leftarrow \mu_0$, $s \leftarrow s_0$, xtol $\leftarrow$ xtol$^\dagger$, ftol $\leftarrow$ ftol$^\dagger$, ctol $\leftarrow$ ctol$^\dagger$\\
        \quad $\nabla^2 F \leftarrow \mathbf I$ \\
	\textbf{Optimization: } \\
	\For{\upshape PenaltyIteration $= 1, 2, \cdots$ m       \label{alg_EFOPT_outerloop_start}}{
		\For{\upshape ConvexifyIteration $= 1, 2, \cdots$ n \label{alg_EFOPT_innerloop_start}}{
			\tcc{Construct and solve TRS}
			$ F(\mathbf x) = $ EvaluateMeritFunction($f, g, h, \mu$) in (\ref{e_merit_func}) \\
			$\tilde F(\mathbf x) = $ ConvexifyProblem($F, \nabla F, \nabla^2 F$) in (\ref{e_quadratic_model}) \\
			$\mathbf d \leftarrow $ Solve TRS in (\ref{e_trs}) by Steihaug-CG algorithm \\
			\tcc{Verify update condition}
			TrueImprove $= F(\mathbf x + \mathbf d) - F(\mathbf x)$ \\
			TRSImprove  $= \tilde F(\mathbf x + \mathbf d) - \tilde F(\mathbf x)$ \\
			ApproxQuality $= $ TrueImprove / TRSImprove \\
			\If{\upshape TrueImprove $< 0$ }{
				Update $\mathbf x \leftarrow \mathbf x + \mathbf d$ \\
				Approximate Hessian $\nabla^2 F \leftarrow$ BFGS algorithm \\
			}
			\tcc{Update trust region}		
			\uIf{\upshape ApproxQuality $< \lambda_1$}{
				Shrink TR: $s \leftarrow \tau^- * s $
			}
			\ElseIf{\upshape ApproxQuality $> \lambda_2$}{
				Expand TR: $s \leftarrow \tau^+ * s$
			}
                \tcc{Convergence test}
			\If{\upshape $\|\mathbf d\| <$ xtol or ${\rm abs(TrueImprove)} < $  ftol}{	
				break
			}		
		} \label{alg_EFOPT_innerloop_end}
            \tcc{Switch convergent thresholds} 
			\If{\upshape C $<$ ctol}{
				xtol $\leftarrow$ xtol$^{\dagger\dagger}$, ftol $\leftarrow$ ftol$^{\dagger\dagger}$, ctol $\leftarrow$ ctol$^{\dagger\dagger}$
			}
		\tcc{Increase penalty coefficient } 
		\eIf{\upshape C $>$ ctol}{
			Increase penalty: $\mu \leftarrow k * \mu$ \\
			Reset Hessian $\nabla^2 F \leftarrow \mathbf I$ \\
		}{
			return $\mathbf x$ \\
		}
	}\label{alg_EFOPT_outerloop_end}
\end{algorithm}

The implementation of EFOPT is outlined in Algorithm \ref{alg_EFOPT}, with assumption that gradients of the objective $f$ and constraints $g,h$ are provided. Initially (Line 18-20), the penalty coefficient $\mu$, trust region size $s$ and approximated Hessian $\nabla^2 F$ are  initialized, while convergence thresholds xtol and ftol are set to  coarse values. Then the solver performs a double loop iteration: the outer loop (Line \ref{alg_EFOPT_outerloop_start}-\ref{alg_EFOPT_outerloop_end}) gradually enforces the optimization constraints by iteratively increasing the penalty weight $\mu$ in the merit function $F(\mathbf x)$, and the inner loop (Line \ref{alg_EFOPT_innerloop_start}-\ref{alg_EFOPT_innerloop_end}) minimizes the merit function $F(\mathbf x)$ by sequentially approximating it locally at each iteration $\mathbf x$ by a TRS problem. Specifically, in the inner loop, the merit function is first evaluated and convexified into quadratic models (Line 24-25). A TRS in (\ref{e_trs}) is then constructed to minimize the approximated merit function subject to trust region, which is solved efficiently by Steihaug's conjugate gradient (Steihaug-CG) method \cite{nocedal1999numerical} (Line 26). Steihaug-CG is known for computational efficiency and has been widely used for trust region sub-problems. The resulting update $\mathbf d$ is evaluated on improvements of the merit function $F$ and approximation quality, which is the ratio of actual and approximated decrease of the merit function (Line 27-29). If the merit function decreases (i.e., improves), the solution vector $\mathbf x$ is updated and the Hessian matrix is estimated using the BFGS algorithm \cite{fletcher2000practical}. After that, the trust region is adjusted for the next iteration: if the approximation quality is poor due to adverse or small improvement in merit function, the trust region size is reduced (Line 34-35), which helps to improve the approximation quality in the next iteration of the inner loop; if the approximation quality is sufficient, the trust region expands for larger updates (Line 36-37); otherwise, the trust region remains unchanged. This adaptive approach, known as Marquardt's strategy, is widely utilized in the Levenberg–Marquardt (LM) algorithm \cite{marquardt1963algorithm}. The inner loop optimization is terminated when the update $\mathbf d$ or merit function variation are below the convergence thresholds xtol and ftol (Line 39-41).  The thresholds are initially set to coarse values xtol$^{\dagger}$, ftol$^{\dagger}$ to avoid the successive optimization in inner loop optimization, which could be time consuming. The reason is that, at the beginning of the outer loop optimization, exhaustive optimization of the inner loop is not necessary since the constraints penalty $\mu$ is small and the exact optimal solution of the merit function is likely infeasible anyway. Once the constraints violation is below a coarse threshold ctol$^{\dagger}$ (Line 43-45), the convergence thresholds of the inner loop switch to smaller values xtol$^{\dagger\dagger}$, ftol$^{\dagger\dagger}$ for greater accuracy. When the inner loop converges or reaches the maximum convexification iteration, the outer loop  assesses the constraint satisfaction. If any constraint is not satisfied (i.e., above the fine threshold ctol$^{\dagger\dagger}$), the penalty weight $\mu$ is increased to intensify the constraint penalty in the merit function (Line 46-48). Otherwise, the optimization returns the optimal result and terminates (Line 49-51). The outer loop repeats until either constraint satisfaction or the maximum penalty iteration is reached, which means the optimization fails.

\begin{table}[t!]
	\centering
	\captionof{table}{Primary difference between EFOPT and TRAJOPT.}
	\begin{threeparttable} 
		\begin{tabular}{l l l }
			\toprule [1 pt]
			Difference  & EFOPT  & TRAJOPT  \\
			\hline
			Convergence threshold  & Adaptive & Fixed \\
			Hessian Approximation  &  BFGS  &  Finite Difference   \\
			Update Condition & True Improvement   & Model Improvement  \\
			TRS Solver &  Steihaug-CG & Gurobi  \\
			TR update strategy & Marquardt's strategy  & Bang-bang strategy \\
			\toprule [1 pt]
		\end{tabular}
	\end{threeparttable} 
	\label{tab_EFOPT_trajopt}
\end{table} 

Compared to TRAJOPT \cite{schulman2014motion}, which is also developed on $\ell_1$ penalty method with SQP, EFOPT exhibits superior computational efficiency and accuracy. There are differences between EFOPT and TRAJOPT discernible from both the article \cite{schulman2014motion} and its source code. Readers are referred to these materials for more detailed information. We summarize four main differences that we believe significantly influence optimization performance in TABLE \ref{tab_EFOPT_trajopt} and discussed as follows.
\begin{itemize}
	\item [1)] EFOPT uses adaptive convergence thresholds -- coarse thresholds to reduce iteration when the solution is far from the feasible region, and fine thresholds to improve the accuracy of feasible solutions. In contrast, TRAJOPT uses a single set of convergence thresholds that may result in either slow convergence or inaccurate solutions at each inner loop optimization.
	
	\item [2)] EFOPT adopts the BFGS algorithm to efficiently approximate the Hessian matrix  $\nabla^2 F$, while TRAJOPT relies on time-consuming finite differences to calculate $\nabla^2  f, \nabla^2  g, \nabla^2  h$ numerically.
	
	\item [3)] In EFOPT, the decision on the current solution update is the true improvement of the merit function.  Conversely, TRAJOPT directly updates $\mathbf x$ to the solution $\mathbf x$ of TRS. Although $\mathbf x$ is optimal for the  quadratic approximation, it does not guarantee to improve the original merit function in case of strong nonlinearity or inappropriate trust region sizes. 
	
	\item[4)] EFOPT uses an efficient C++ implementation of Steihaug's conjugate gradient algorithm to solve trust-region sub-problems, while TRAJOPT employs a commercial  solver Gurobi\footnote{https://www.gurobi.com/} which is designed for general QPs. Repeatedly deploying Gurobi at each inner loop iteration leads to a certain overhead in time consumption. 
	
	\item[5)] EFPOT utilizes Marquardt' strategy to shrink, expand or maintain the trust region size based on the approximation quality, while TRAJOPT' strategy either shrinks or expands the trust region. The bang-bang  strategy in TRAJOPT tends to cause fluctuations, resulting in longer convergence or failures in practice.
\end{itemize}

These differences could lead to significant discrepancies in performance in trajectory optimization, which is evident in our benchmark results presented in Section \ref{sec_benchmark}.

\section{System Overview}
\label{sec_sys_overview}
\begin{figure*}[bthp!] 
	\centering
	\includegraphics[width=1\linewidth]{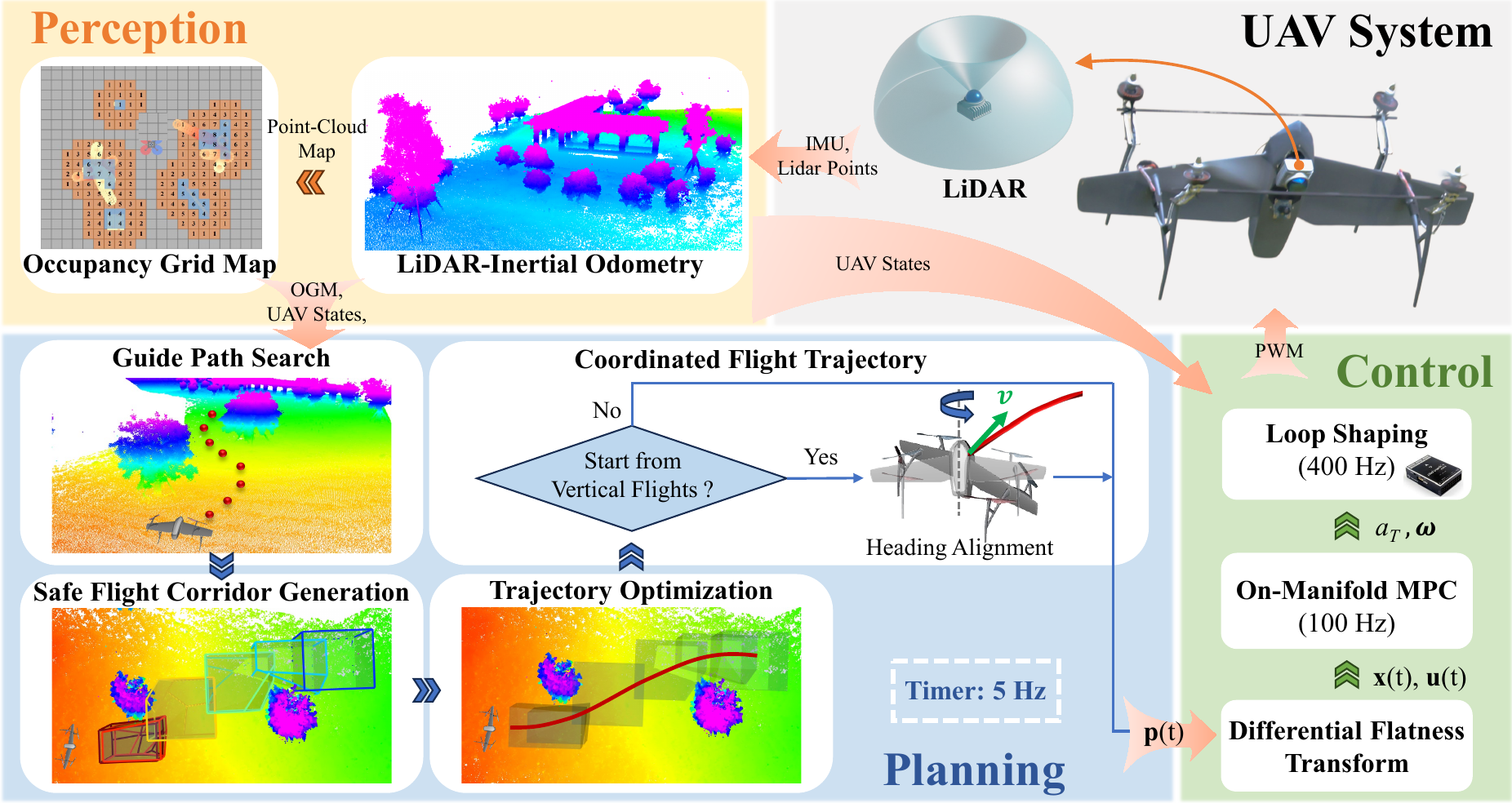} 
	\caption{System overview} 
	\label{fig_system_overview}
\end{figure*}
In this section, the algorithm pipeline for autonomous tail-sitter \textcolor{black}{in both simulations and real-word experiments} is presented. The framework consists of three main modules: perception, planning and control, as depicted in Fig. \ref{fig_system_overview}. Within the perception module, vehicle states including pose, velocity, and attitude, are estimated by a LiDAR-inertial odometry \cite{xu2022fast}, using 3-D LiDAR points and IMU measurements. Simultaneously, a high-accuracy 3-D point-cloud map is constructed and continuously updated in real time. This map is then downsampled and utilized to maintain a local occupancy grid map (OGM), segmenting the environment into spaces that are obstacle-occupied, unknown and known-free, as proposed in \cite{ren2023rog}. 

Next, the planning module generates a desired trajectory based on the OGM and current UAV states. Initially, a guide path is searched between the current UAV position and the target within a replanning horizon (set to \SI{30}{m} in this paper) in know-free and unknown spaces, using A* algorithm. Once successfully finding a collision-free path, a safe flight corridor (SFC) comprising of consecutively overlapped polyhedra is generated along the guide path, to segment obstacle-free space from the map. This SFC is then incorporated as a positional constraint in trajectory optimization. Due to the limited sensing range of the onboard sensor and receding horizon, the planning module consisting of the A* path search, SFC generation and trajectory optimization runs at a fixed frequency of \SI{5}{Hz}, to avoid newly-appeared obstacles. If the initial state involves vertical flights (i.e., hovering, takeoff and landing), the current UAV heading may not be aligned with the velocity direction on the trajectory. In this case, an extra yawing trajectory is inserted at the beginning of the optimized trajectory to align the UAV current heading with the initial velocity direction of the trajectory. With the heading alignment, the tail-sitter can perform smooth coordinated flights on the trajectory.  While the A* path search and SFC generation are implemented following existing works \cite{hart1968formal, liu2017planning} respectively, our primary focus in this paper is on trajectory optimization.

The planned trajectory $\mathbf{p}(t)$ is then forwarded to the control module as a reference. In this module, the trajectory planned in flat-output space is converted into a state-input trajectory $(\mathbf x(t), \mathbf u(t))$ by the differential flatness transform in \cite{lu2024trajectory}. Using the current state feedback from LIO, an on-manifold model predictive control (MPC) proposed in \cite{lu2022manifold, lu2024trajectory} tracks the state-input trajectory at a frequency of \SI{100}{Hz}. The consequent optimal thrust acceleration command $a_T$ and body angular velocity commands $\boldsymbol{\omega}$ are sent to an autopilot, where they are tracked by a low-level controller employing loop-shaping techniques (i.e., proportional–integral–derivative (PID) controller and additional Notch filters for flexible modes attenuation \cite{xu2019full}) at \SI{400}{Hz}. The low-level controller then generates throttle and torque commands, which are further converted into pulse width modulation (PWM) signals to drive the actuators (i.e., four motors).

Except \textcolor{black}{for} the low-level controller implemented on autopilot, all other modules -- perception, planning and control modules -- run in real time on the onboard computer.

\begin{figure*}[t!] 
	\centering
	\includegraphics[width=1\linewidth]{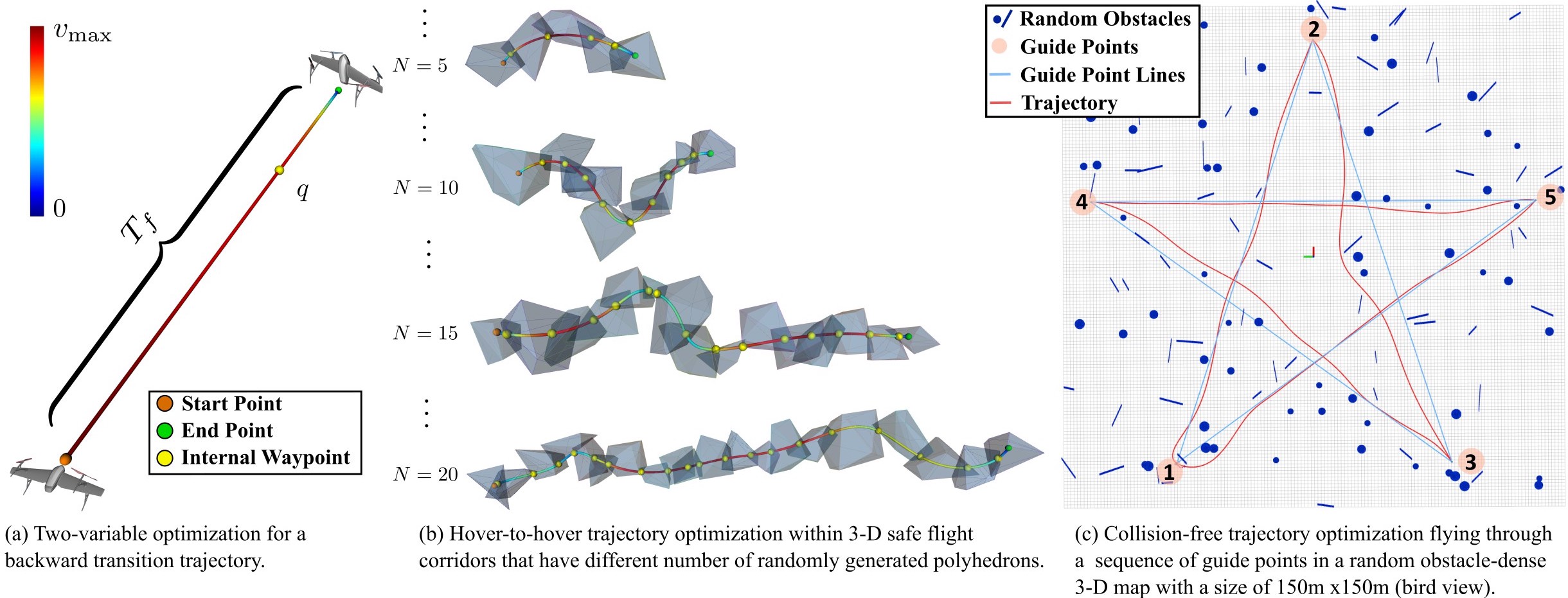} 
	\caption{Three benchmark simulations to evaluate the performance of different solvers in trajectory optimization.} 
	\label{fig_sim}
\end{figure*}

\section{Benchmark Results}
\label{sec_benchmark}
In this section, we conduct three simulated trajectory optimization to benchmark the performance of our proposed feasibility-assured solver, EFOPT, against off-the-shelf NLP solvers, including TRAJOPT \cite{schulman2014motion}, \textcolor{black}{KNITRO \cite{knitro}} SNOPT \cite{snopt}, IPOPT \cite{ipopt}, NLOPT \cite{nlopt} and LBFGS-Lite\footnote{https://github.com/ZJU-FAST-Lab/LBFGS-Lite}. Fig. \ref{fig_sim} depicts the three simulations with detailed explanation provided later. 
  

In all three simulations, the convergence thresholds of all solvers are set ${\rm xtol} = 1\mathrm{e}{-6}$ and ${\rm ftol} = 1\mathrm{e}{-6}$. Constraint tolerance is set ${\rm ctol} = 1\mathrm{e}{-6}$ in the first and second simulations, and relaxed to ${\rm ctol} = 1\mathrm{e}{-3}$ in the third simulation for efficiency consideration in online planning. These thresholds are used as the fine thresholds in EFOPT to ensure a fair comparison. In addition, the coarse thresholds in EFOPT are ${\rm xtol}^{\dagger\dagger} = 1\mathrm{e}{-3}$, ${\rm ftol}^{\dagger\dagger} = 1\mathrm{e}{-3}$ and ${\rm ctol}^{\dagger\dagger} = 1\mathrm{e}{-3}$ for fast convergence.  The first two simulations are conducted on a 5.3 GHz Intel i9 processor, while the last one is carried out on the tail-sitter onboard computer with a 1.8 GHz Intel i7 processor.

\begingroup
\color{black}
To enhance versatility and robustness to numerous types of NLPs, solvers typically offer multiple algorithm options and available parameters. This flexibility allows to achieve the most effective use of a package. However, solver's performance can be significantly affected by its settings. Therefore, determining configurations that yield optimal performance on benchmark problems, is essential for a fair comparison. First, we select proper algorithms or third-party packages for internal computation. We choose the ``Sequential Least-Squares Quadratic Programming" method in NLOPT, as it is the only available option for constrained NLPs. In KNITRO,  we select the``Interior-Point" method that outperforms the ``Active-Set" method in tests. We opt for the "MA97" package as the linear solver in IPOPT instead of the default  ``MUMPS" package due to its generally better performance.  Second,  setting limits of iteration and computational time is crucial for balancing success rates with computational efficiency. Initially, we set these limits as large as possible in simulation 2, then gradually reduce them until success rates begin to decrease significantly. As a result, we set maximum iteration counts of 10,000 for SNOPT, 3,000 for TRAJOPT and KNITRO, and 256 for LBFGS-Lite, while setting a time limit of 5 seconds for both IPOPT and NLOPT. Third, we supply analytical gradients of the objective function and constraints to improve convergence speed. Gradients have been verified by the derivative check function of SNOPT and KNITRO, but this function is disabled during tests to reduce time consumption. Besides, the default setup of TRAJOPT yields a low success rate, so we adopt the trust-region parameters from EFOPT to effectively enhance its performance.  Unlike the other solvers, LBFGS-Lite cannot handle constraints directly. Instead, it solves unconstrained NLPs as defined in (\ref{e_merit_sqp}) with a fixed penalty weight $\mu$ that must be specified manually. Hence we evaluate its performance across three different $\mu$ values in simulations. The parameters and convergence thresholds outlined above are commonly used settings for regular implementation. 
\endgroup

%

\begin{figure}[t!] 
	\centering 
	\subfigure[The solutions of different solvers. The white area denotes the feasible region, where the solutions have constraint violation less than tolerant ${\rm ctol} = 1\mathrm{e}{-6}$. Green markers denote feasible solutions, while orange markers denote infeasible solutions that have constraint violation exceeding the tolerance.]{   
		\includegraphics[width=0.95\linewidth]{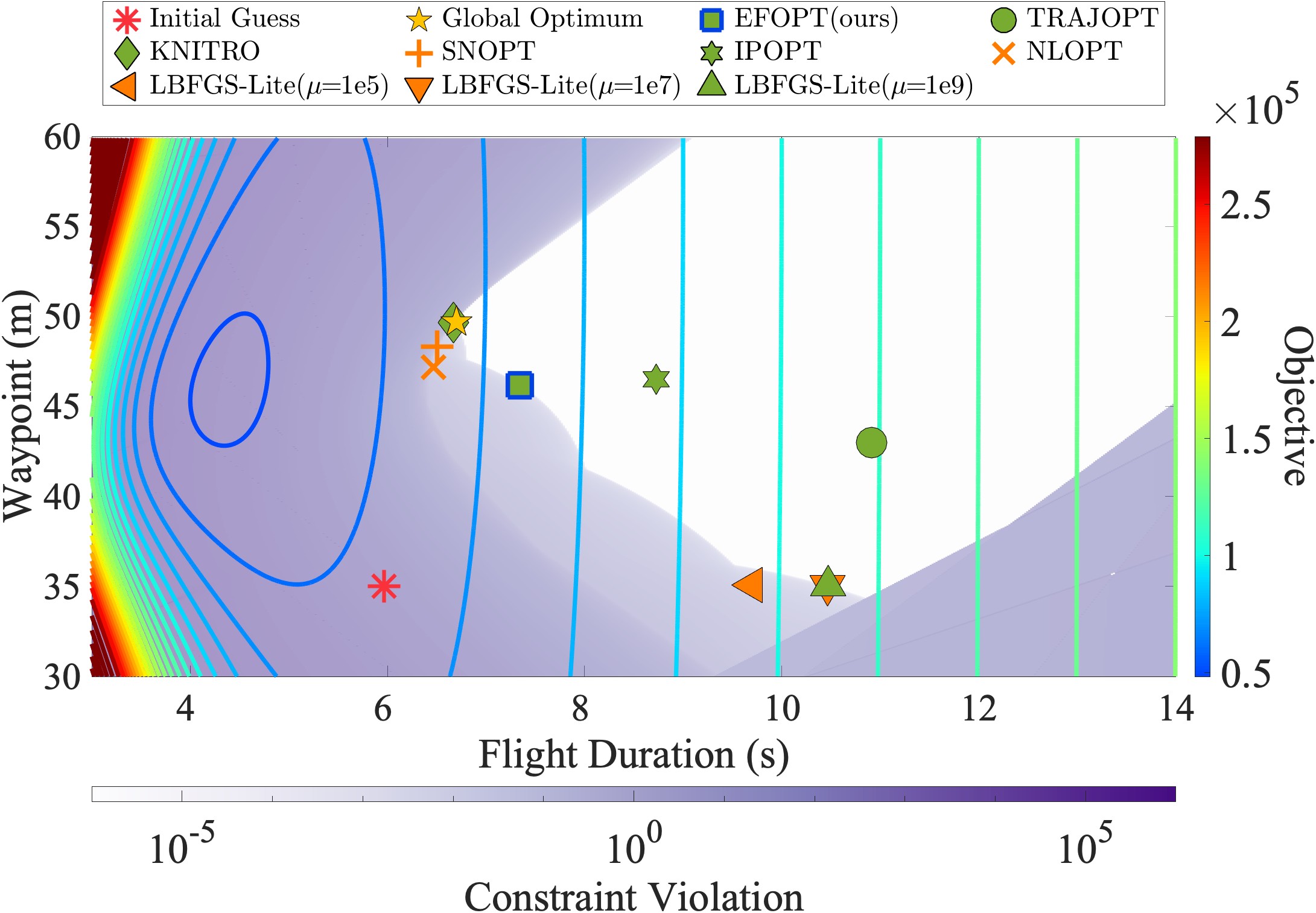} 
		\label{fig_bm1_2dim_result}    
	}
	\subfigure[Convergence path of  EFOPT and LBFGS-Lite$(\mu = 1\mathrm{e}{9})$.]{   
		\includegraphics[width=0.95\linewidth]{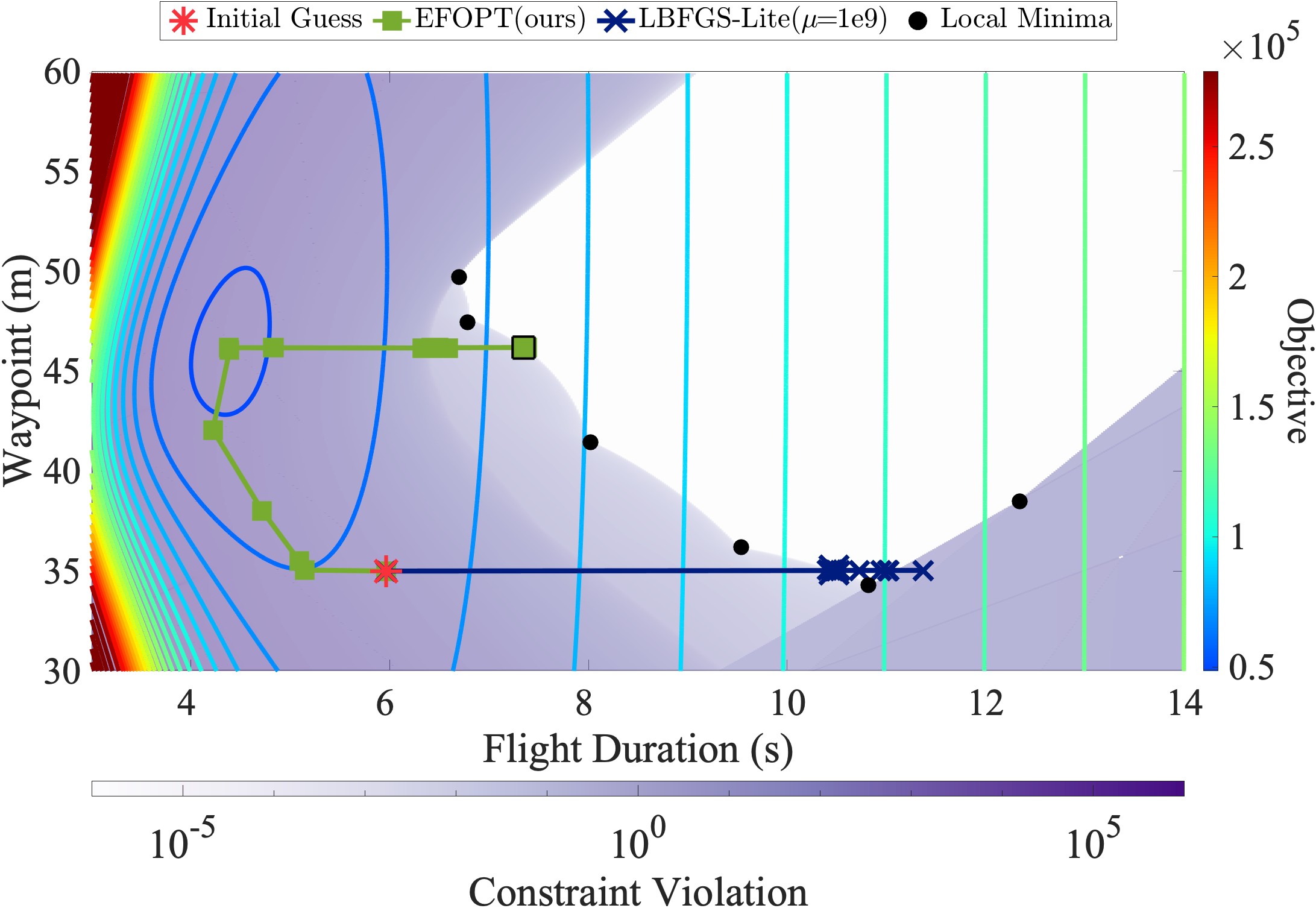}  
		\label{fig_bm1_2dim_path}   
	}
	\caption{Convergence progress in the two-variable optimization for the backward transition trajectory in Fig. \ref{fig_sim} (a).} 
	\label{fig_bm1_2dim}
\end{figure}

\subsection{Two-variable optimization}
\label{sec_bm1}
Visualizing the change of objective function versus the values of optimization variables at different iterations can provide an intuitive presentation of solver convergence behavior. Such visualization is possible when the number of optimization variables is less than or equal to two. Therefore, in this simulation, we consider a typical tail-sitter backward transition flight and formulate a two-variable optimization problem for its trajectory generation. As shown in Fig. \ref{fig_sim}(a), Given an initial state $\mathbf s_0$ in level fight at \SI{15}{m/s} and a terminal state $\mathbf s_f$ in hovering, the transition flight is a horizontal 2-D straight-line trajectory parameterized by two scalar variables: the total flight duration $T_f$ and an internal waypoint $ q$ restricted on the straight line. Each trajectory segment separated by the waypoint $q$ accounts for half of the total flight time $T_f$, leading to a  two-segment polynomial trajectory $\mathbf C(q, T_f, \mathbf s_0,  \mathbf s_f)$ as defined in (\ref{e_poly}). The initial values of the two optimization variables $T_f$ and $q$ are  chosen as heuristic values $(p_f -  p_0) / v_{\rm max}$ and $(p_0 + p_f) /2$, respectively.

\begin{table}[t!]
	\centering
	\captionof{table}{Solvers' performance in simulation 1.}
	\begin{threeparttable} 
		\begin{tabular}{l l l l }
			\toprule [1 pt]
			Solver  & Objective(${\times 10^4}$)  & Constraint & Time (ms)  \\
			\hline
			\textbf{EFOPT}  &  7.3  &  $\textbf{0}$ & \textbf{8.8}  \\
			TRAJOPT &  10.9 & 3.6$\times 10^{-7}$ & 230.7 \\
			\textcolor{black}{KNITRO} & 6.7 & 7.1$\times 10^{-10}$ &  157.9\\
			SNOPT &  \textbf{6.5} &  0.10 & 42.6 \\	
			IPOPT &  8.7 &  \textbf{0} & 210.3 \\
			NLOPT & \textbf{6.5} &  0.14 & 353.1 \\
			LBFGS-Lite{\scriptsize$ (\mu \!=\! {\rm 1e5})$} &  9.7 &  0.05  &  9.3  \\
			LBFGS-Lite{\scriptsize$ (\mu \!=\! {\rm 1e7})$} &  10.5 &  1.5$\times 10^{-6}$  &  9.9  \\
			LBFGS-Lite{\scriptsize$ (\mu \!=\! {\rm 1e9})$} &  10.5 &  2.9$\times 10^{-9}$  &  14.4  \\
			\toprule [1 pt]
		\end{tabular}
	\end{threeparttable} 
	\label{tab_bm1_result}
\end{table} 

In Fig. \ref{fig_bm1_2dim}, we visualize the objective values and constraint violations of the two-variable trajectory optimization, as well as the results of different solvers in comparison. The objective values are depicted by contours in varying colors, while the degree of constraint violations is indicated by the depth of purple color. It is seen from the contours and color map that both the objective and constraint are nonlinear. The white area where constraint violations are less than the tolerant ${\rm ctol} = 1\mathrm{e}{6}$, indicates the feasible region. Notably, the objective has a global minimum outside the feasible region. The initial guess marked by a red asterisk is infeasible and distant from the global minimum of objective. Fig. \ref{fig_bm1_2dim_result} shows the global optimum (i.e., a feasible solution with the lowest objective value, as indicated by a yellow star) and solutions of all compared solvers.  \textcolor{black}{TABLE \ref{tab_bm1_result} summarizes solution values and computational time. It is seen that SNOPT, NLOPT and LBFGS-Lite with lower penalty weights (i.e., $(\mu = 1\mathrm{e}{5}, 1\mathrm{e}{7}$), whose solutions are marked in orange, yield infeasible solutions in the purple area. Solutions of EFOPT, KNITRO, TRAJOPT, IPOPT and LBFGS-Lite$(\mu=1\mathrm{e}{9})$, are marked in green, denoted feasible solutions have been found. Among the feasible solutions, KNITRO  achieves the global optimum, while EFOPT and LBFGS-Lite$(\mu=1\mathrm{e}{9})$ converge on the feasible boundary. TRAJOPT and IPOPT terminate within the feasible region, where the objective can still decrease without constraint violation in any direction. EFOPT has the fastest computation time among all eight solvers and the second lowest objective value among five feasible solutions. Although KNITRO finds the global optimum with lower objective cost than EFOPT,  KNITRO takes 157.9 ms, which is  almost 20 times slower than EFOPT. From the perspective of online trajectory optimization, EFOPT is more reliable in practice that it efficiently finds a feasible solution with a slight loss of optimality.}


Fig. \ref{fig_bm1_2dim_path} presents the convergence paths of EFOPT and LBFGS-Lite$(\mu=1\mathrm{e}{9})$ towards feasible solutions. In EFOPT, the penalty weight $\mu$ starts from a small value ($\mu = 1$ in this paper), which makes the objective dominate in the initial merit function in (\ref{e_merit_sqp}). Hence EFOPT first converges towards the objective global minimum  along contour gradients, and then the convergence direction shifts towards the feasible region on the right. EFOPT finally terminates at a point on the boundary of the feasible region, without further converging to even nearby local minima that are also feasible {(the black dots in Fig. \ref{fig_bm1_2dim_path})}. The reason is that the terminal penalty weight has been increased to $\mu = 1\mathrm{e}{7}$. Such a large weight ensures feasibility but meanwhile overwhelms the gradient direction of the merit function. The resultant gradient is barely affected by the objective, hence preventing a gradient-based solver (e.g., EFOPT) from further decreasing the objective. On the other hand,  LBFGS-Lite ($\mu = 1\mathrm{e}{9}$) uses a large, fixed penalty weight of $\mu = 1\mathrm{e}{9}$, which drives the solution straightly to the feasible region along the constraint gradients. After reaching the feasible region, the solver moves back along the constraint gradient to achieve lower objective values. Eventually, the solver converged at the boundary of the feasible region without further converging to even its nearby local minima (black dots in Fig. \ref{fig_bm1_2dim_path}), a phenomenon similar to our EFOPT due to the same reason. It is noted that the EFOPT reaches termination with a smaller penalty weight of $\mu = 1\mathrm{e}{7}$,  while LBFGS-Lite with the same penalty weight fails to converge to a feasible solution. In practice, a larger weight would increase numerical difficulties that may result in failures or slow convergence, which is evident in Fig. \ref{fig_bm1_2dim_path} that LBFGS-Lite spends plenty of iterations near the convergence result.


\begin{remark}
	\rm
	Both EFOPT and TRAJOPT are based on $\ell_1$ penalty method and SQP. However, in this test, TRAJOPT yields an inferior solution, where the objective value can further decline in any direction without violating any constraints. We believe the reason is that TRAJOPT directly uses the TRS solution as the current update, without evaluating the improvement of the original merit function. Besides, we have attempted to enable SNOPT to find feasible solutions by increasing the iteration numbers but failed. Similar findings have been reported in \cite{sun2021fast}, where  SNOPT often terminates without convergence or fails to find a feasible solution.
\end{remark}

\subsection{Random safe flight corridor}
\begin{figure}[t!]
	\centering
	\includegraphics[width=1\linewidth]{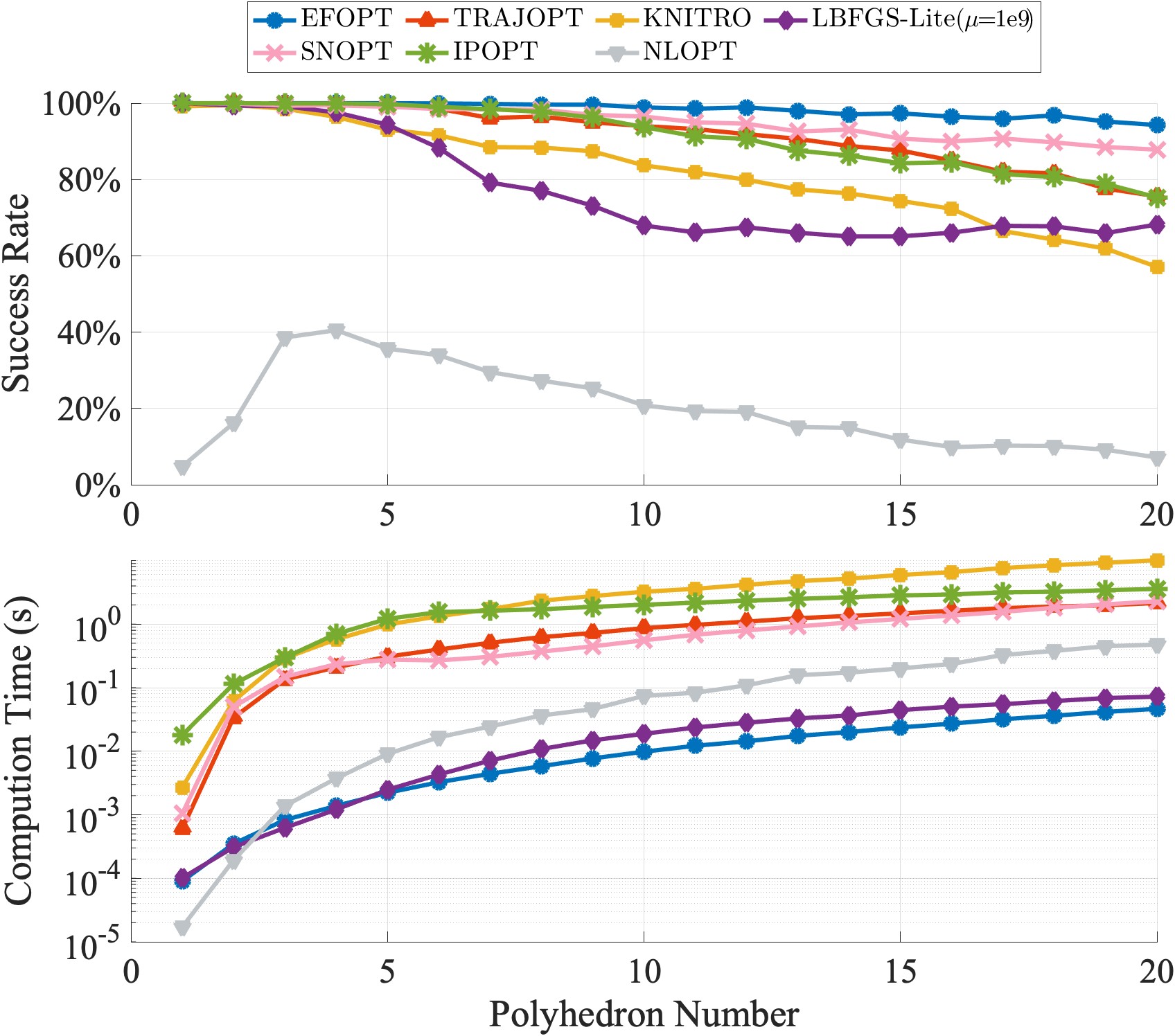} 
	\caption{Success rate and average computation time solvers in hover-to-hover trajectory optimization within randomly generated 3-D SFCs in Fig. \ref{fig_sim} (b).} 
	\label{fig_bm2_sfc}
\end{figure}
In the second simulation, we benchmark solvers in the problem of trajectory optimization with different scales of SFC constraints (Fig. \ref{fig_sim}(b)). A SFC consists of $N$ overlapping polyhedra with $N$ tested from 1 to 20. For each $N$, we generate 1,000 different SFCs by randomly specifying the face number (in the range of 6 to 24), shape, and position of each polyhedron in the SFC. This results in 20,000 tested SFCs in total, among which the longest SFC spans more than \SI{100}{m}. The corresponding number of optimization variables when optimizing trajectories in these SFCs ranges from 1 to over 470, thereby testing the solvers' performance in both small-scale and large-scale scenarios.  To further increase the variable dimension, we use a barycentric coordinate to represent waytpoint  $\mathbf q_i$, which should be constrained within the intersected polyhedra $\mathcal P_i \cap \mathcal P_{i+1}$, by the vertices of the intersected polyhedra as proposed in \cite{wang2022geometrically}. That is, $\mathbf q_i =  f_{\mathcal B} (\boldsymbol{\xi}_i)$ with $\boldsymbol{\xi}_i$ the vertices. Hence, The optimization variables become $(\boldsymbol{\xi}, \mathbf T)$.  For each waypoint $\mathbf q_i$, there are often many vertices representing the intersected polyhedra,  hence the dimension of $\boldsymbol{\xi}_i$ will be significantly higher than that of $\mathbf q_i$, which effectively increases the scale of the optimization problem. The initial conditions of $\mathbf q_i$ (or its representation $\boldsymbol{\xi}_i$) are set to the mean of vertices of each overlapped polyhedra. Time allocation $\mathbf T$ is heuristically initialized as $ t_i = \|\mathbf q_i - \mathbf q_{i-1}\| / v_{\rm max}$.
  
Fig. \ref{fig_bm2_sfc} presents the success rate and average computation time for optimizing trajectories in the 20,000 SFCs. Among the seven solvers under evaluation, EFOPT demonstrates the highest success rate across all polyhedron numbers, and maintains the lowest computation time for polyhedron numbers from 5 to 20. As the polyhedron number increases, EFOPT's success rate slightly decreases from $100\%$ to $94.3\%$, and its time consumption rises from \SI{0.1}{ms} to \SI{466}{ms}. LBFGS-Lite $(\mu = 1\mathrm{e}{9})$ requires slightly higher computation time, but its success rate drops significantly when the polyhedron number is above 5, reaching the lowest value of $65\%$ at 15 polyhedra. TRAJOPT, \textcolor{black}{KNITRO}, SNOPT and IPOPT exhibit similar patterns in both success rate and computation time. Their computation times are about 100 times longer than EFOPT for nearly all polyhedron numbers. NLOPT consistently has the lowest success rate for all polyhedron numbers.

\subsection{Obstacle-dense random map}\label{sim:obs_avoidnace}

The third simulation assesses the complete planning module by planning obstacle-free trajectories on a randomly generated, obstacle-dense map measuring $150\times$\SI{150}{m^2}, as shown in Fig. \ref{fig_sim}(c). We specify a sequence of five points (i.e., the guide points) to roughly guide the UAV to fly a pentagram-shaped trajectory. At each replan cycle, the planning module takes the current guide point in the list as its intermediate local target and generates a smooth trajectory from the current UAV position to the target, using the pipeline of A*, SFC generation, and trajectory optimization detailed in Section \ref{sec_sys_overview}. Once the current guide point is reached (the UAV is \textcolor{black}{within \SI{2}{m}}), it will be removed from the list and the next guide point will be used in the next replan cycle. The replan frequency is set to \SI{5}{Hz}, allowing \SI{200}{ms} for the planning module to complete. 
If the planner \textcolor{black}{times out} and or the planned trajectory violates any constraint,  the UAV continues to follow the previous trajectory. Following the previous trajectory at a planning failure could lead to collisions with obstacles. If it occurs, we reset the UAV to the position immediately before the collision and continue the replan process as usual. The success rate is calculated by the ratio of feasible optimization results (regardless of its time consumption) to the total number of trajectory optimizations. In this simulation, the optimization variables $(\mathbf Q, \mathbf T)$ are initialized in the same way as in simulation (b).

\begin{figure}[bpht!]
	\centering
	\includegraphics[width=1\linewidth]{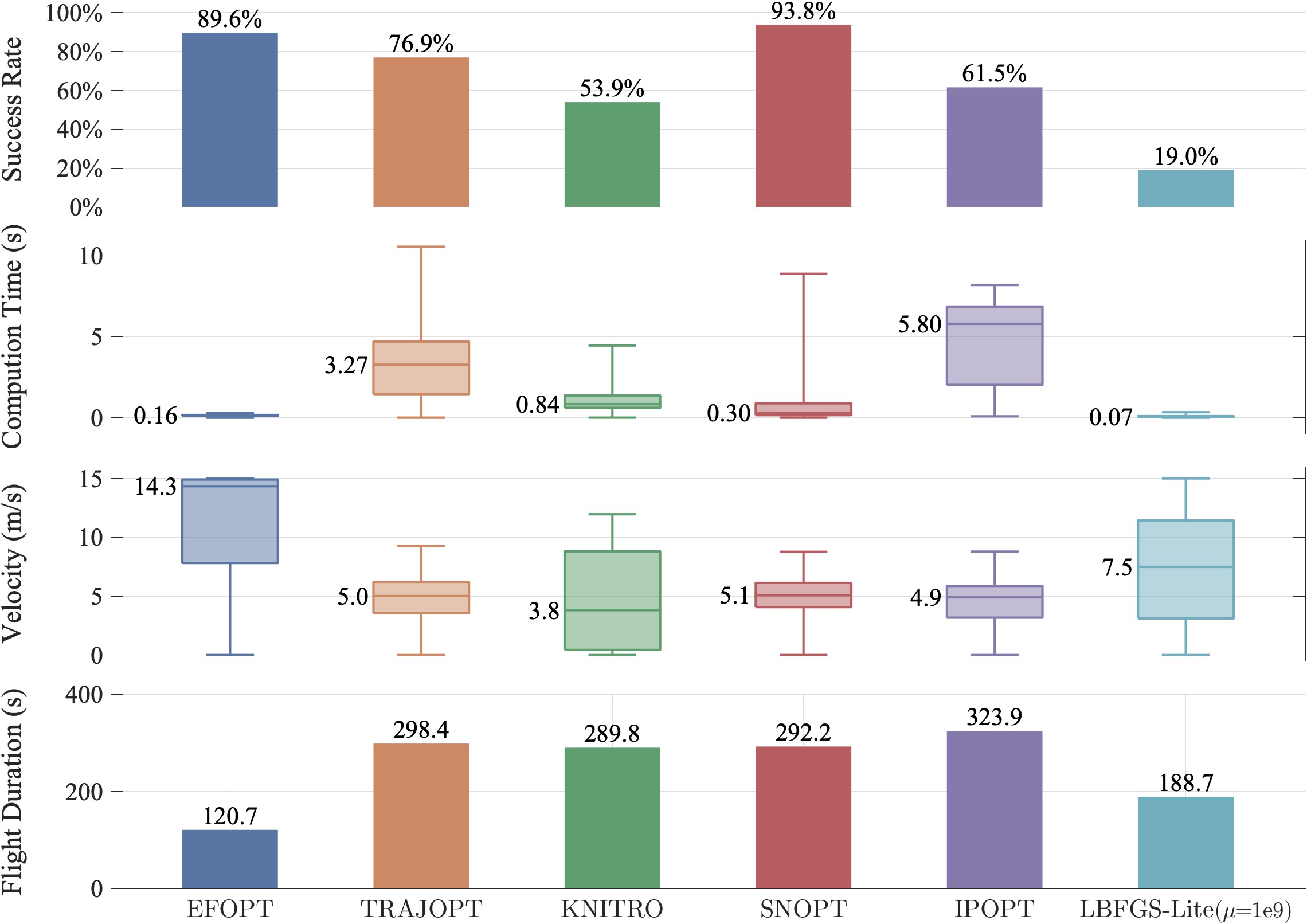} 
	\caption{Benchmark results of local replanning in a 3-D environment with random obstacles as in Fig. \ref{fig_sim} (c).} 
	\label{fig_bm3_randomMap}
\end{figure}
\begin{figure}[bpht!]
	\centering
	\includegraphics[width=1\linewidth]{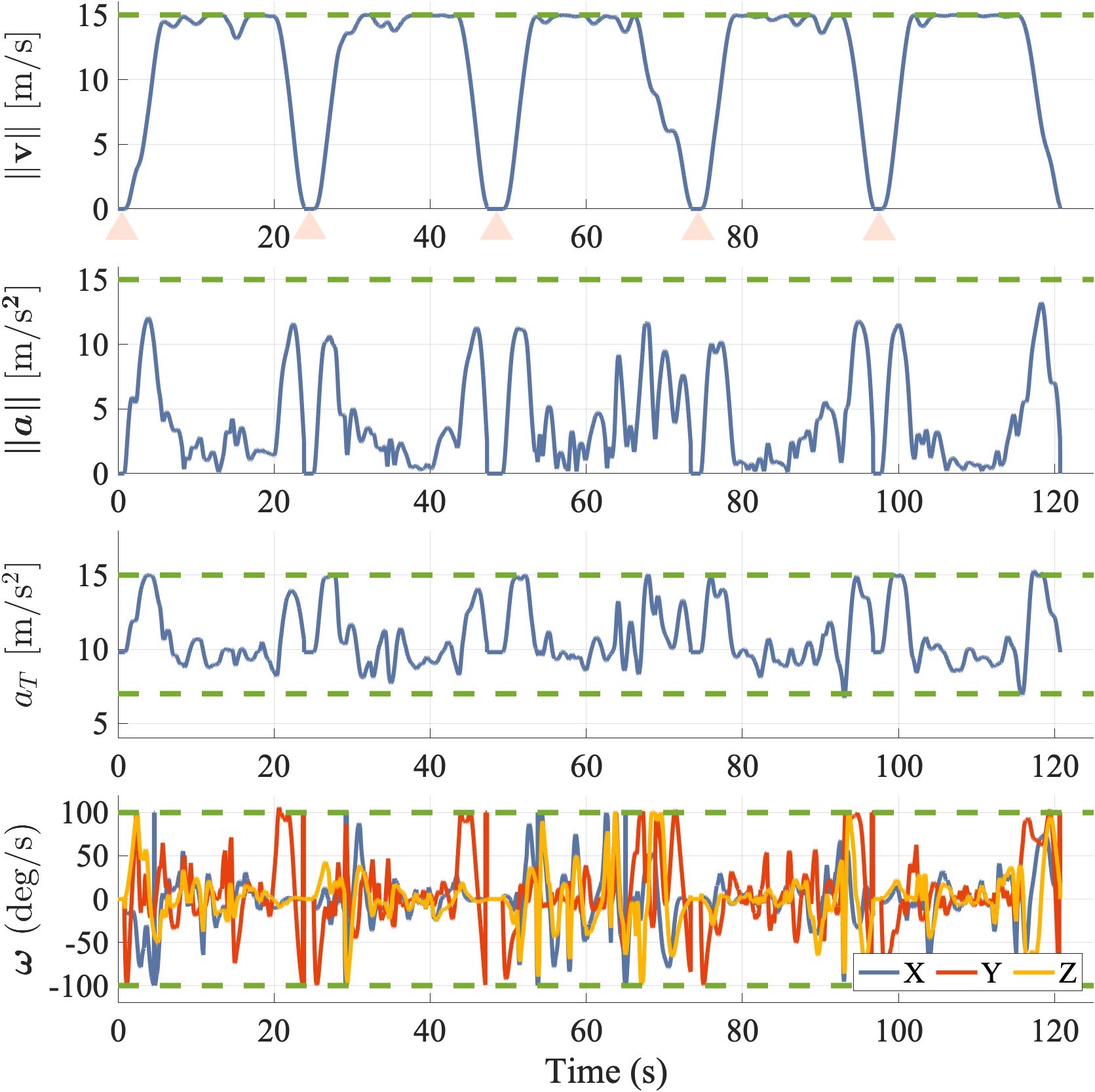} 
	\caption{UAV states and inputs of a trajectory solved by EFOPT in the simulation problem in Fig. \ref{fig_sim} (c). The pink triangles in the first figure denote the time when the UAV arrives at the guide points and executes heading alignment toward the next guide point. } 
	\label{fig_randomMap}
\end{figure}

Fig. \ref{fig_bm3_randomMap} shows the planning performance of the benchmarked solvers except NLOPT, which was unable to complete the test. As can be seen, EFOPT achieves the second highest success rate with a small gap from the highest one SNOPT. Although SNOPT presents the highest success rate, its time consumption (i.e., \SI{0.3}{s} and \SI{8.8}{s} in median and maximum)  indicates that it frequently fails to update trajectory at \SI{5}{Hz} (i.e., timeout). In case of \textcolor{black}{a} timeout, the UAV slows down or even stops at the end of the previous trajectory. The low flight speed considerably simplifies the optimization problem, which is the main reason for SNOPT's high success rate. 
In terms of computation time, EFOPT and LBFGS-Lite ($\mu = 1\mathrm{e}{-9}$) are the only two methods that can run in real time. Relatively speaking, although LBFGS-Lite ($\mu = 1\mathrm{e}{-9}$) has a lower computation time than EFOPT, its success rate is much lower ($19.0\%$ versus $89.6\%$), causing a bias in its average computation time. 

The high success rate and \textcolor{black}{computational} efficiency of EFOPT have enabled it to plan feasible trajectories timely, even at high speeds. As a consequence, EFOPT achieves the highest flight speed and the shortest flight duration among all evaluated solvers (see the lower two plots in Fig. \ref{fig_bm3_randomMap}). An example trajectory planned by EFOPT is shown in Fig. \ref{fig_randomMap}. As can be seen, EFOPT effectively pushes the vehicle speed to the maximum value of \SI{15}{m/s} for the majority of the flight time, while maintaining compliance with actuator constraints on thrust acceleration $a_T$ and angular velocity $\boldsymbol{\omega}$.

\begin{figure}
	\centering
	\includegraphics[width=0.9\linewidth]{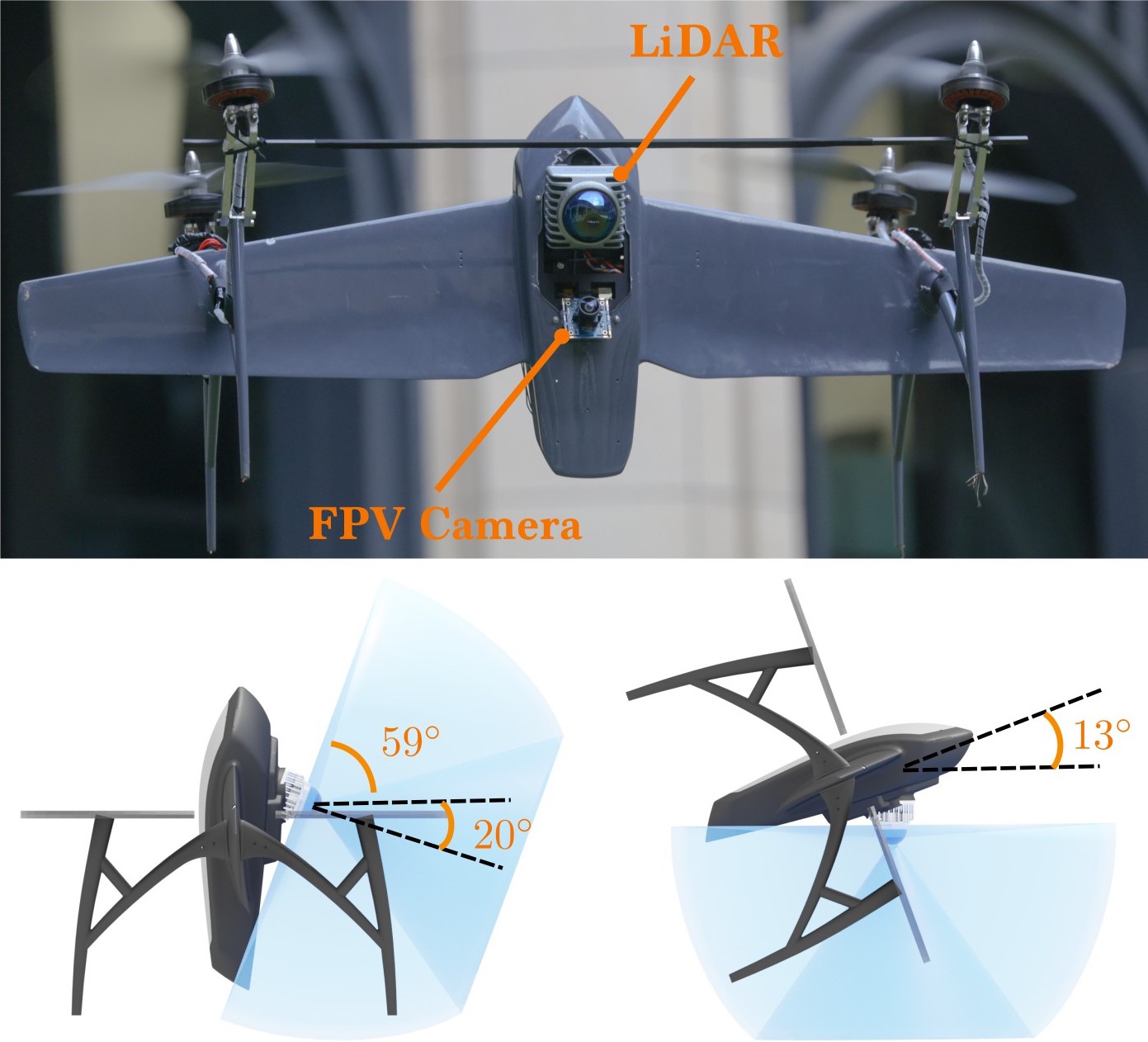} 
	\caption{LiDAR installation on the tail-sitter and its FoV. The LIVOX MID-360 LiDAR is mounted at a tilted angle of $20^\circ$, providing a FoV of $59^\circ \times 360^\circ$.} 
	\label{fig_fov}
\end{figure}

\section{Real World Experiments}
\label{sec_experiment}
\begin{figure*}[!htbp] 
	\centering 
	\subfigure[Indoor drone laboratory.]{   	\includegraphics[width=1\linewidth]{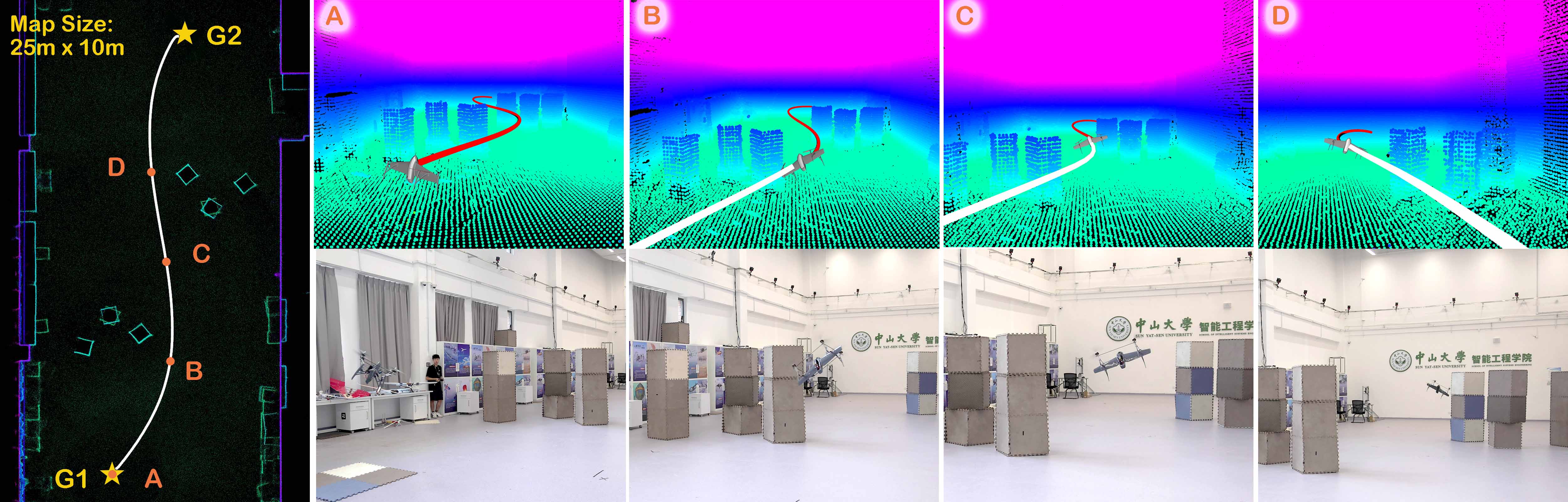} 
		\label{fig_indoor}    
	}
	\subfigure[Underground parking lot. Snapshot images A,B are captured by two fixed cameras, while C and D are captured by an onboard FPV camera.]{   
		\includegraphics[width=1\linewidth]{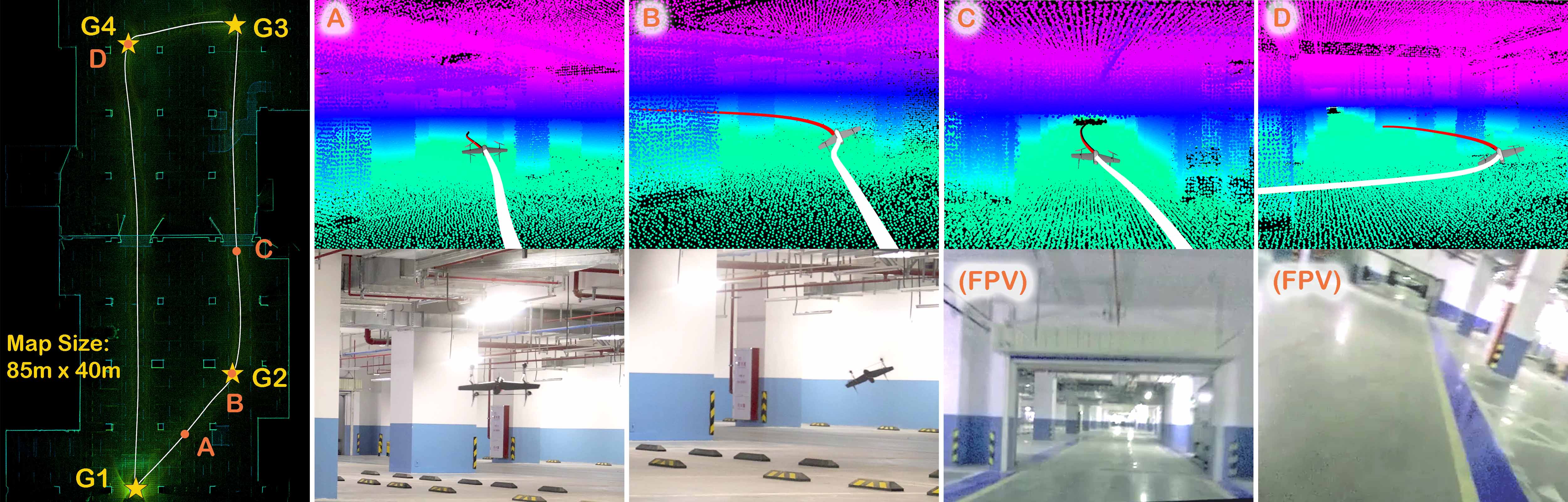}  
		\label{fig_parking}   
	}
	\subfigure[Outdoor park at Sun Yat-sen University. Image C presents the local planner generates a trajectory flying over a pavilion and bushes.]{   
		\includegraphics[width=1\linewidth]{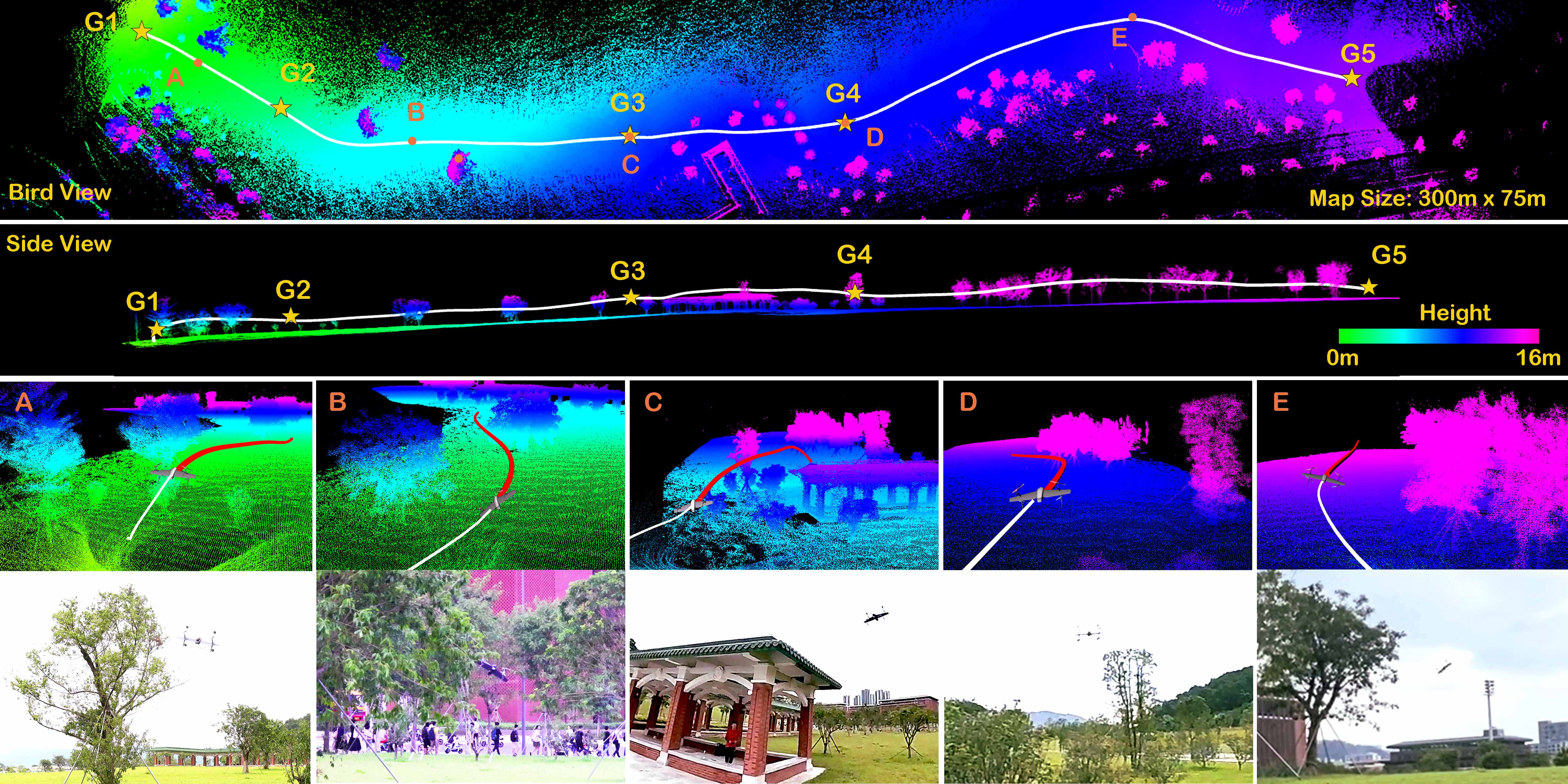}  
		\label{fig_outdoor}   
	}     
	\caption{Autonomous flights in real-world environments. Point-cloud maps are constructed by FAST-LIO2 in real time. Yellow stars (G1-G5) are guide points. Snapshots (A-E) include local trajectories and camera images along the flight trajectories, while their corresponding positions are marked by orange dots on the point-cloud maps.}  
	\label{fig_realworld_exp}
\end{figure*}

In this section, we detail the implementation of an autonomous tail-sitter UAV system, as shown in Fig. \ref{fig_fov}. To validate our entire algorithm framework, we conduct high-speed autonomous flight tests in various real-world environments, including an indoor drone laboratory, an underground parking lot, and an outdoor park at Sun Yat-sen University, as shown in Fig. \ref{fig_realworld_exp}. Guide points, depicted as yellow stars on the point-cloud maps, are provided to roughly guide the shape of the flight path by acting as local goal positions in the replanning module as detailed in Section \ref{sim:obs_avoidnace}. Obstacles are present between two successive guide points, and sharp turns are needed to fly from one guide point to the next. Avoiding such obstacles and ensuring dynamic feasibility of the trajectory are both achieved in the planning module. 

\subsection{Hardware platform}
As shown in Fig. \ref{fig_fov}, our UAV platform is based on a tail-sitter prototype derived from our previous airframe design \cite{gu2018coordinate}. It is equipped with a Livox MID-360 LiDAR\footnote{https://www.livoxtech.com/mid-360},  featuring a FoV of $59^\circ \times 360^\circ$ and a detection range of \SI{40}{m}. The onboard computing is implemented on a DJI Manifold 2-C\footnote{https://www.dji.com/manifold-2/specs} (\SI{1.8}{GHz} quad-core Intel i7 CPU), followed by an autopilot PX4 Mini\footnote{https://px4.io/}. Additionally, a camera is installed solely for first-person-view (FPV) video capturing. The UAV platform has a total weight of \SI{2.7}{kg} and a wingspan of \SI{90}{cm}. As shown in Fig. \ref{fig_fov}, the LiDAR is mounted on the airframe's belly at a tilt angle of $20^\circ$, ensuring continuous forward visibility for the UAV during a transition with pitch variation of $72^\circ-13^\circ$, which covers most maneuvers from vertical to level flights of the UAV. 

\begin{figure*}
	\centering
	\subfigure[Indoor drone laboratory.]{   
		        \includegraphics[width=1\linewidth]{
		        		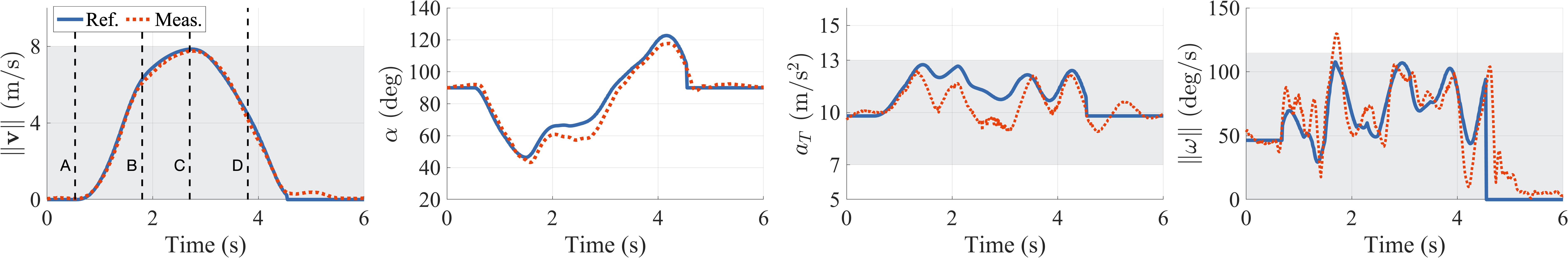}  
		        \label{fig_realworld_indoor}   
		    }
	\subfigure[Underground parking lot.]{   
		\includegraphics[width=1\linewidth]{
			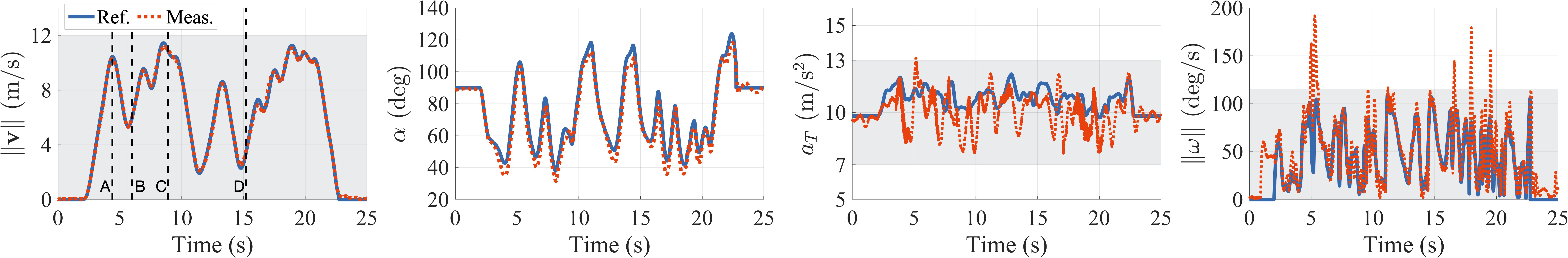}  
		\label{fig_realworld_parking}   
	}
	\subfigure[Outdoor park.]{   
		\includegraphics[width=1\linewidth]{
			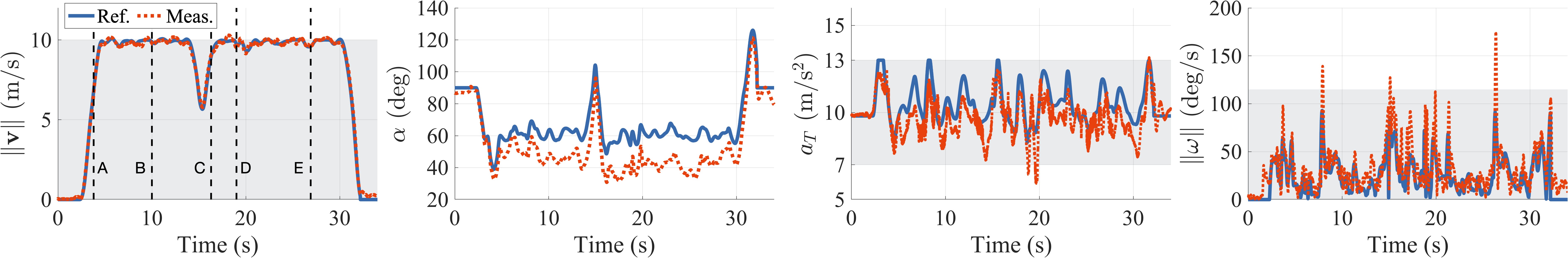}  
		\label{fig_realworld_outdoor1}   
	}
	\subfigure[Outdoor park at high speeds.]{   
		\includegraphics[width=1\linewidth]{
			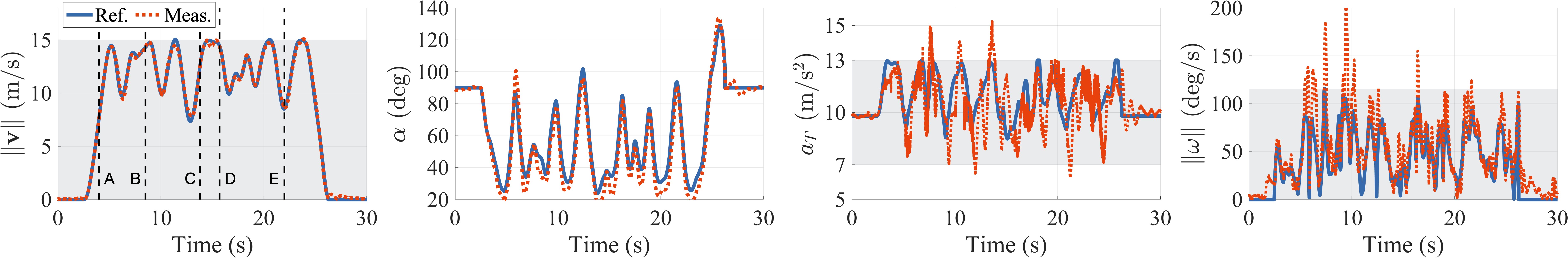}  
		\label{fig_realworld_outdoor2}   
	}
	\caption{Flight data of real-world experiments. Shaded regions denote constraint boundaries in the trajectory optimization. Black dashed lines in figures in the first column denote the time stamps of the snapshots in Fig. \ref{fig_realworld_exp}.}
\end{figure*}

\subsection{Indoor drone laboratory}
 In the first flight experiment as shown in Fig. \ref{fig_indoor}, we place obstacles in an indoor flight space measuring $25\times 10$\SI{}{m}$^2$. As marked by yellow stars G1 and G2, two guide points are used, which are also the start and terminal positions of the UAV. Two stacks of blocks are placed between the initial and terminal positions, preventing the UAV from a straight-line flight. Initially, the UAV hovers at the start position G1 and detects the obstacles in front. It plans a trajectory through the free space on the right side. Because the initial Yaw angle does not match the angle derived from the differential flatness transform on the trajectory, the tail-sitter UAV did not immediately execute a coordinated flight on the trajectory, but instead adjusted its Yaw angle to align with the trajectory's tangent first. Then the vehicle successfully flies a S-shape trajectory while avoiding obstacles, and arrives at the terminal position G2.  Readers are encouraged to watch the video demonstration for the heading alignment behavior and the complete flights. 
 
  Fig. \ref{fig_realworld_indoor} presents the flight data.  It is seen that the planned reference velocity quickly reaches the maximum of \SI{8}{m/s} in around two seconds, while the angle of attack (AoA) $\alpha$ decreases from $90^\circ$ to $40^\circ$.  The optimized trajectory then pulls up the AoA $\alpha$ over $120^\circ$ to perform  aggressive deceleration, and finally reaches hovering status.  During the flight, the reference velocity and  control inputs of thrust acceleration $a_T$ and angular velocity $\boldsymbol{\omega}$ are confined within the constraint boundaries (i.e., the shaded regions), ensuring the trajectory is feasible to be tracked by controllers and actuators. In consequence, the measurements of vehicle states (e.g., velocity and AoA) and inputs (e.g., thrust acceleration and angular velocity) track their reference  accurately under the  on-manifold MPC and low-level controllers.

\subsection{Underground parking lot}
In the second flight experiment, the autonomous tail-sitter UAV flies through a more challenging underground parking lot, as shown in Fig. \ref{fig_parking}. We designated the start and terminal position as the same point G1, and set three other guide points G2-G4 at the turning corners, guiding the UAV to fly a loop trajectory. It is seen that the parking lot presented numerous obstacles, including dense pillars and complex pipes on the ceiling, which was about \SI{5}{m} high. Additionally, two gates at points C and its left side limit the passage height to under \SI{3}{m}. The flight path involves consecutive sharp turns requiring dynamically feasible control inputs and consistently low passage height confining the safe space, a problem posing significant challenges to precise perception, planning, and control. As captured in the snapshots in Fig. \ref{fig_parking}, the tail-sitter transitions swiftly to level flight from the start position to point A, executes a banked turn at B, and successfully passes through the first gate at C by flying a straight-line trajectory. After passing through the gate at C, the UAV makes two consecutive sharp $90^\circ$ turns at points G3 and G4 (D). It then flies through the second gate and returns to the origin position G1. 

As shown in Fig. \ref{fig_realworld_parking}, the UAV velocity experiences significant variations during the \SI{25}{s} flight, due to the complex environment and the aggressive sharp turns. The  velocity peaks at \SI{11.4}{m/s}, slightly below the constraint bound of \SI{12}{m/s}.  The UAV decelerates to \SI{2}{m/s} twice for  sharp turns at \SI{11.5}{s} and \SI{14.8}{s}, because of the sharp turns at G3 and G4. After each sharp turn, the vehicle quickly accelerates above \SI{8}{m/s}. It is also seen that the AoA $\alpha$ also has large fluctuations ranging from $35^\circ$ to $120^\circ$ for frequent acceleration and deceleration. Despite the challenging environment and aggressive flight maneuvers, the optimal trajectory successfully restricts the control input $a_T$ and $\boldsymbol{\omega}$ within the constraint boundaries.  Fluctuations of measured control inputs are more severe due to the MPC execution for precise control of velocity and AoA.

\subsection{Outdoor park}   
In the third flight experiment, we verify the system in a large-scale outdoor environment -- a park measuring $300\times 75$\SI{}{m}$^2$, as shown in Fig. \ref{fig_outdoor}. We set start and terminal positions G1 and G5, along with three guide points G2-G4 as depicted by yellow stars, to roughly guide the flight paths. This flight path cannot guarantee obstacle avoidance, as it is seen that trees are blocking the path between each two guide points of G2-G5. There is also a pavilion between G3 and G4. Moreover, the park's terrain, a slope rising about \SI{16}{m} from left to right, is evident from the varying colors of the point cloud. Therefore, the tail-sitter UAV is required to fly through the obstacles with terrain variation, posing a significant navigation challenge. The flight trajectory (the white curve in Fig. \ref{fig_outdoor}) shows that the tail-sitter safely flies \textcolor{black}{through} the free space beside trees at points A and B, and climbs up to fly over the pavilion and bushes between G3 and G4. The UAV then detects thick and tall trees between G4 and G5, thus it flies along the edge of these trees from point D to E, and finally arrives at termination G5. 

We conduct flight tests at two different maximum velocities of \SI{10}{m/s} and \SI{15}{m/s}, as shown in the flight data in Fig. \ref{fig_realworld_outdoor1} and \ref{fig_realworld_outdoor2}. In the lower speed test, the vehicle consistently maintains the maximum speed at \SI{10}{m/s} for most of the time, except for a slowdown to \SI{6}{m/s} at \SI{15}{s} right before point C. At that moment, the UAV performs an aggressive climb at height by pulling up the pitch angle (see the equivalent increase in AoA $\alpha$), preparing to fly over the pavilion and bushes between C and D, as shown in Fig. \ref{fig_outdoor}.  Similar to the previous two flight tests,  constraints on control inputs of $a_T$ and $\boldsymbol{\omega}$ are fully satisfied, resulting in accurate tracking in velocity.  However, there is a steady-state error around $15^\circ$ in the AoA $\alpha$, as shown in Fig. \ref{fig_realworld_outdoor1}. We suspect this error is attributed to the wind disturbance. 
In spite of AoA errors, the MPC is still robust to stabilize the UAV and tracks the velocity accurately.

 In the high-speed test as shown in Fig. \ref{fig_realworld_outdoor2}, the UAV frequently reaches the maximum velocity of \SI{15}{m/s}, but cannot maintain this speed due to the complex environments and actuator limitations. The AoA experiences more significant variations and fluctuations than the lower speed test, to perform aggressive acceleration and deceleration. The minimum and maximum AoA reach $20^\circ$ and $130^\circ$, fully exploiting the FoV limitation of LiDAR and flight envelope. The optimal control inputs are also constrained within the required boundaries.  Although the measured control executions and AoA variation are apparently increased,  the system still presents accurate  tracking performance in velocity and AoA, owing to the dynamically feasible trajectory optimization and real-time MPC controllers. 

 \subsection{Computation time}

\begin{figure}[t!]
	\centering
	\includegraphics[width=1\linewidth]{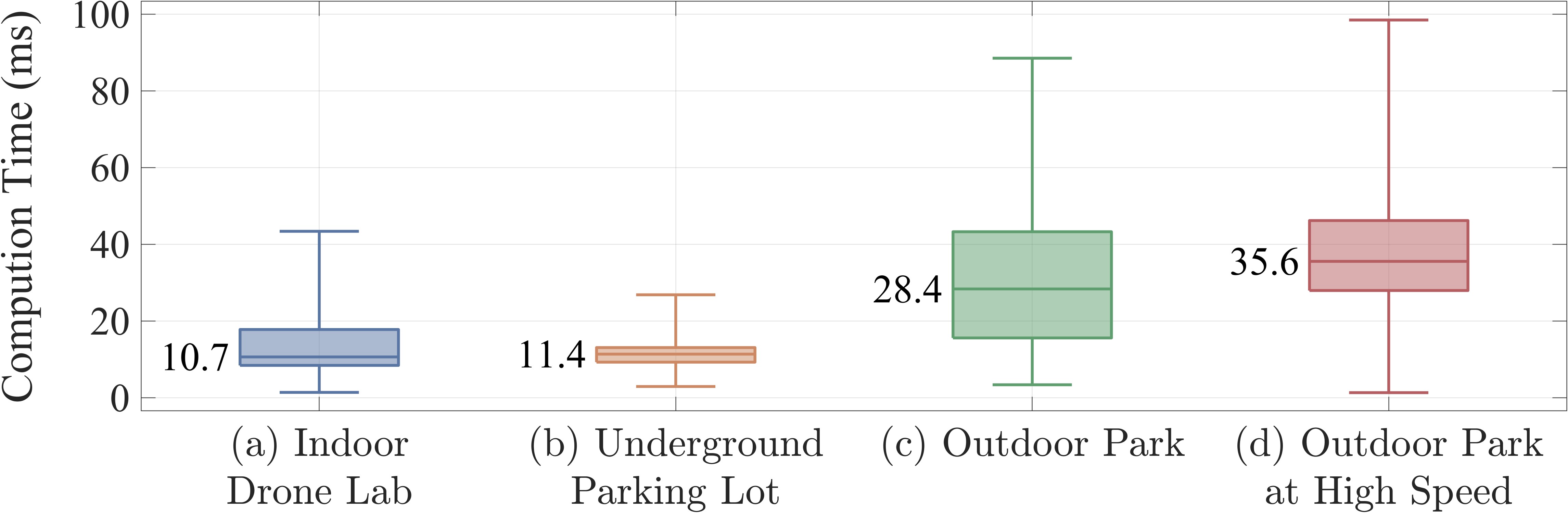} 
	\caption{Computation time of trajectory optimization in real-world experiments.} 
	\label{fig_realworld_time}
\end{figure}

{The time consumption of trajectory optimizations in these four real-world flight tests is presented in Fig. \ref{fig_realworld_time}. The median computation time rises from \SI{10.7}{ms} to \SI{35.6}{ms} along test (a) to (d), along with the increasing  environmental complexity and trajectory aggressiveness. The maximum time to solve a trajectory optimization arrives at \SI{98.5}{ms} in the \textcolor{black}{fourth} test, but is still less than \textcolor{black}{half} the planning interval of \SI{200}{ms}, which is sufficient for real-time planning at \SI{5}{Hz}. 
}

\section{Discussion}
\label{sec_discuss}
In this section, we discuss the limitations and extensions of the proposed system.
\subsection{LiDAR selection and installation}
The LIVOX MID-360 LiDAR, chosen for its light weight, wide FoV, and capability to capture point clouds from a considerable distance of \SI{40}{m}, is integral to our prototype's design for autonomous flights. It offers the best thrust-to-weight ratio among the lightest commercial LiDAR sensors. However, as shown in Fig. \ref{fig_system_overview}, this sensor has a blind zone at its top. The current installation prioritizes continuous forward FoV, crucial for obstacle detection and avoidance during transition and level flights, at the cost of sacrificing visibility along the belly. This setup, while effective for our intended demonstrations, is not optimal for applications like geographic surveying and infrastructure inspection, where terrain mapping beneath the airframe is important. In the future, the development of  lightwight and wide-FoV LiDAR sensors specifically tailored for UAV applications, could expand the practicality and applicability of our system in real-world scenarios.

\subsection{LiDAR SLAM}
LiDAR-based SLAM, leveraging IMU data and point cloud scans, provides accurate and low-latency state estimation as well as dense 3-D maps. However, LiDAR-based SLAM approaches easily fail in environments with few available geometric features, such as open spaces and narrow tunnels. This presents a challenge for tail-sitter UAVs designed to execute multi-scale missions across diverse environments, ranging from cluttered environments to open space at high altitudes, where LiDAR's effectiveness diminishes. In this paper, our focus is on trajectory generation with obstacle avoidance for autonomous flights, thus we particularly consider obstacle-dense environments where the deployed LIO, FAST-LIO2, can perform reliably. Moreover, we found a gap in real-time multiple-sensor fusion and SLAM research that integrates LiDAR, global navigation satellite system (GNSS) and  visual cameras. Such integration could be more robust for diverse environments,  and potentially beneficial for VTOL UAV applications.

\subsection{Uncoordinated flights}
We consider a coordinated flight trajectory generation framework in this paper. \textcolor{black}{The} coordinated flight condition offers benefits in terms of aerodynamic efficiency, moment stability, and model reduction. However, from a practical perspective, there are specific scenarios, such as sideways flights through narrow gaps, where uncoordinated flights at low speeds may be necessary. The differential flatness transform utilized in our approach supports the specification of an additional yaw angle, in singularity conditions where the airspeed is zero $\|\mathbf v_a\| = 0$. This allows low-speed uncoordinated flights by incorporating extra yaw planning. Hence, the proposed framework can be extended to accommodate such specialized flight maneuvers, enhancing its applicability in diverse operational scenarios.

\subsection{Safe trajectory planning}
Our proposed trajectory generation framework plans trajectories in both known-free and unknown spaces. However, there are scenarios where trajectory safety may not be fully guaranteed. For example, unobserved hidden obstacles could exist within the safe flight corridors in complex unknown spaces, and their sudden detection could significantly complicate the trajectory optimization process. In cases where the optimization fails, the reference trajectory may not be updated, potentially leading to collisions. These challenges have been addressed in \cite{tordesillas2019faster}, which suggests planning an additional safe trajectory exclusively in the known-free space. If the primary trajectory optimization fails, the system can switch to this safe trajectory which is absolutely obstacle-free. We aim to incorporate this safety strategy into future tail-sitter planning frameworks, but adding an extra trajectory planning layer would introduce a considerable computational burden.

\subsection{EFOPT for general NLPs}
The proposed EFOPT solver, characterized by its efficient feasibility-assured approach, is formulated in a form of general NLPs. Hence, it is potentially suitable to a wide range of NLP applications where finding a feasible solution is the primary concern. However, it's important to note that the current version of EFOPT is tailored for relatively small-scale optimizations. It relies on dense matrix calculations using the Eigen library\footnote{https://eigen.tuxfamily.org}. As a result, the current EFOPT is more suitable to NLPs involving dozens of variables. Future versions could be adapted to handle larger-scale problems, broadening its applicability.

\section{Conclusion}
\label{sec_conclusion}
We have proposed a fully autonomous tail-sitter UAV capable of high-speed flights in unknown, cluttered environments. The hardware platform and algorithm framework integrated with perception, planning and control technologies, are presented in detail. \textcolor{black}{We propose an optimization-based planning framework, providing with collision-free and dynamically-feasible trajectories for autonomous tail-sitter navigation at high speeds. We also address challenges of solving constrained, non-convex trajectory optimization in real-time, by developing an efficient, feasibility-assured optimizer, EFOPT, which enhances computational efficiency and ensures feasibility by relaxing optimality to a slight extent.} The performance of EFOPT are compared with various state-of-the-art NLP solvers through simulations, demonstrating a superior computational efficiency and feasibility satisfaction in the context of tail-sitter trajectory optimizations. Extensive flight experiments in diverse real-world environments provide convincing evidence of the proposed tail-sitter UAV's effectiveness, indicating the system's promising potential for a broad spectrum of practical applications.

\section*{Acknowledgment}
This work was supported in part by Hong Kong RGC ECS under grant 27202219 and in part by DJI donation.

\appendix
\subsection{Flight dynamics}
\label{app_dynamics}
\begin{figure}[t!] 
	\centering
	\includegraphics[width=0.85\linewidth]{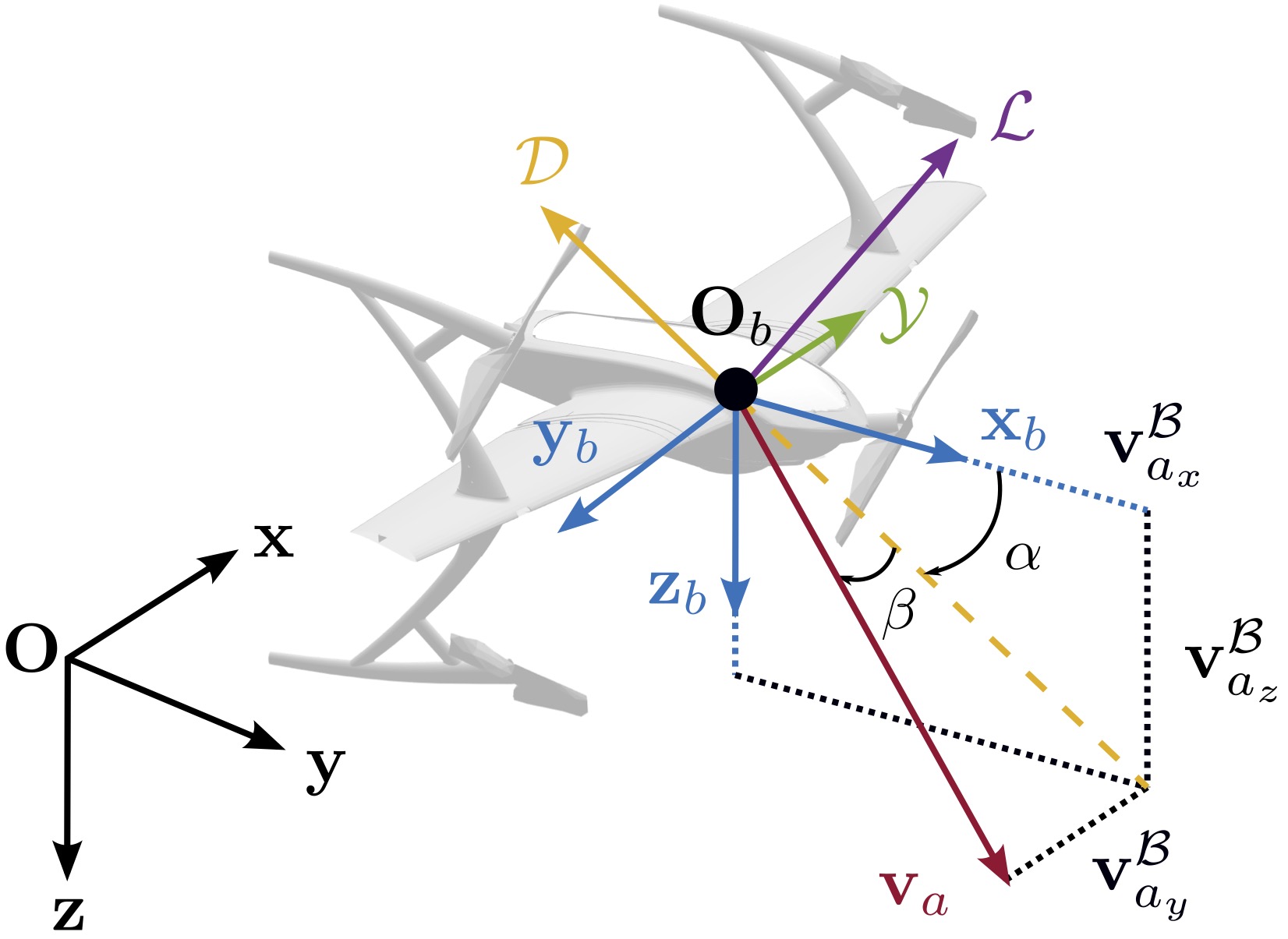} 
	\caption{Coordinate systems and aerodynamic nomenclatures.} 
	\label{fig_flight_dyn}
\end{figure}

\begin{figure}[t!] 
	\centering
	\includegraphics[width=0.85\linewidth]{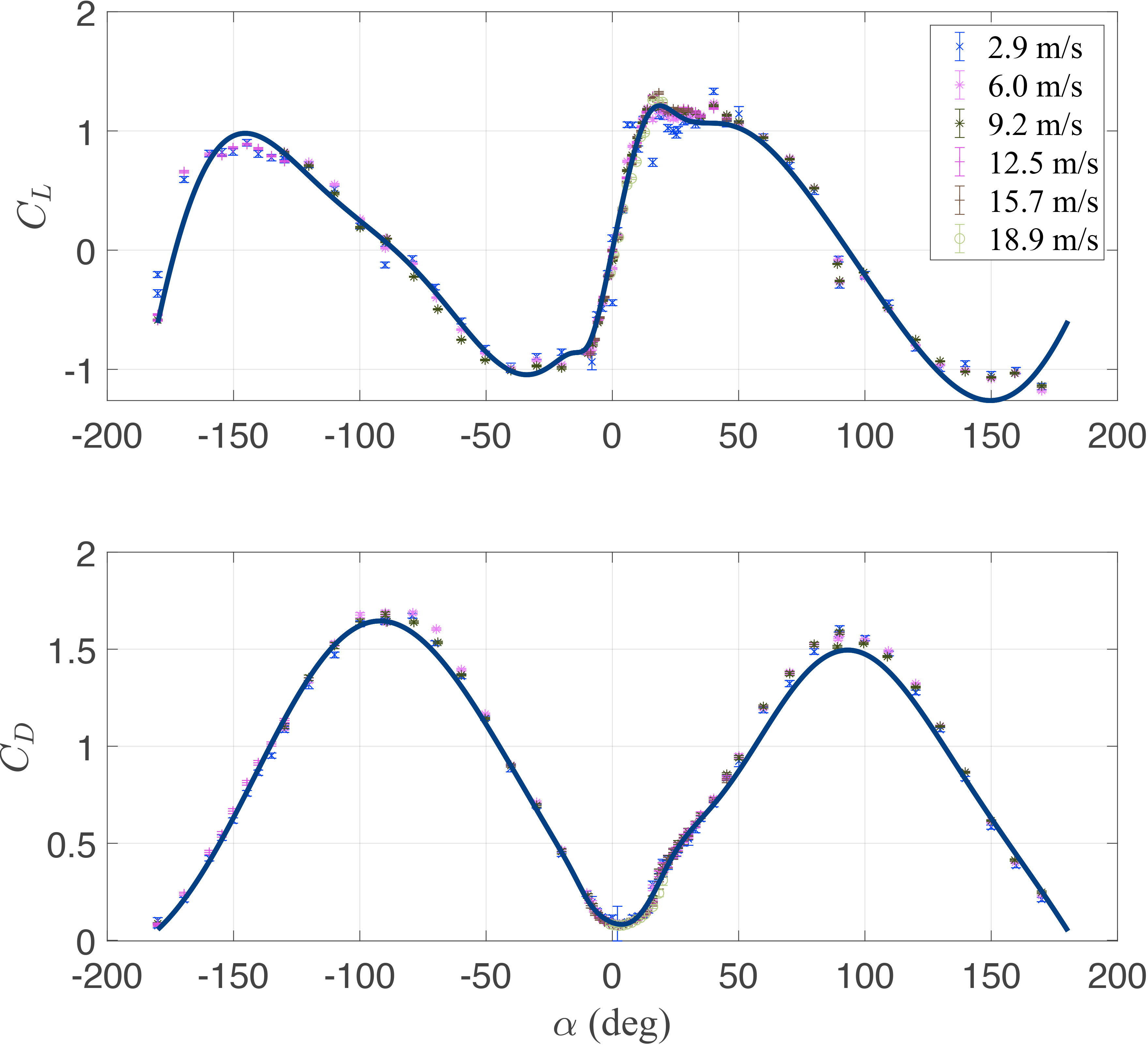} 
	\caption{Longitudinal aerodynamic coefficients $C_L$ and  $C_D$ of our previous quadrotor tail-sitter UAV prototype, identified by wind tunnel tests \cite{lyu2018simulation}.} 
	\label{fig_aero_coef}
\end{figure}

The coordinate frames and flight dynamics are defined and  presented in Fig. \ref{fig_flight_dyn}, following the convention of traditional fixed-wing aircraft. We view the tail-sitter airframe as a rigid body. The translational and rotational dynamics of the aircraft are modeled as follows:
\begin{subequations}
	\label{e_vehicle_dyn}
	\begin{align}
		\dot{\mathbf{p}} &= \mathbf{v} \label{e_translation_kin} \\
		\dot{\mathbf{v}} &= \mathbf{g} + a_T\mathbf{R}\mathbf{e}_1 + \frac{1}{m}\mathbf{R}\mathbf{f}_a  \label{e_translation_dyn} \\
		\dot{\mathbf{R}} &= \mathbf{R}\lfloor\boldsymbol{\omega}\rfloor \label{e_rotation_kin} \\
		\mathbf{J}\dot{\boldsymbol{\omega}} &= \boldsymbol{\tau} + \mathbf{M}_a - \boldsymbol{\omega} \times \mathbf{J} \boldsymbol{\omega} 
		\label{e_rotational_dyn}
	\end{align}
\end{subequations}
where $\mathbf{p}$ and $\mathbf{v}$ are respectively the vehicle position and velocity in the inertial frame, $\boldsymbol{\omega}$ is the angular velocity in the body frame, $\mathbf{R}$ denotes the rotation from the inertial frame to the body frame, $m$ is the total mass of the aircraft, $\mathbf{J}$ is the inertia matrix and $\mathbf{g} = [0 \ 0\ 9.8]^T$ is the gravity vector in the inertial frame. $a_T$ and $\boldsymbol{\tau}$ denote the thrust acceleration scalar and control moment vector produced by actuators (e.g., four motors for a quadrotor tail-sitter). $\mathbf{f}_a$ and $\mathbf{M}_a$ are the aerodynamic force and moment in the body frame, respectively. The notation $\lfloor \mathbf \cdot \rfloor$ denotes the skew-symmetric matrix of a 3-D vector.

Referring to \cite{etkin1959dynamics}, the aerodynamic force  $\mathbf{f}_a$ is modeled in the body frame as follows:
\begin{align}
	\label{e_aerodynmics}
	\mathbf{f}_a =  \begin{bmatrix}
		\mathbf f_{a_x} \\
		\mathbf f_{a_y} \\
		\mathbf f_{a_z}
	\end{bmatrix} = \begin{bmatrix}
		-\cos \alpha && 0 && \sin \alpha \\
		0 && 1 && 0 \\
		-\sin \alpha && 0 && -\cos \alpha
	\end{bmatrix} \begin{bmatrix}
		\mathcal{D} \\ \mathcal{Y} \\ \mathcal{L}
	\end{bmatrix}  
\end{align}
where $\alpha$ is the angle of attack. The force components $\mathcal{L}, \mathcal{D}, \mathcal{Y}$ are respectively the  lift, drag, and side force, which  can be written as products of non-dimensional coefficients, dynamic pressure $\frac{1}{2} \rho V^2$, the reference area $S$ (e.g., the wing area) as follows: 
\begin{equation}
	\begin{aligned}
		\mathcal{L} &= \frac{1}{2} \rho V^2 SC_L(\alpha,\beta) \\
		\mathcal{D} &= \frac{1}{2} \rho V^2 SC_D(\alpha,\beta) \\
		\mathcal{Y} &= \frac{1}{2} \rho V^2 SC_Y(\alpha,\beta)  
	\end{aligned} 
	\label{e_aerodyn}
\end{equation}
where $\rho$ is the air density and $V = \| \mathbf v_a \|$ is the norm of the airspeed. $C_L, C_D, C_Y$ are the lift, drag, and side force coefficients. The aerodynamic coefficients are functions of the angle of attack $\alpha$ and the sideslip angle $\beta$, depending on the design of the airfoil profile and the overall airframe. The accurate aerodynamic coefficients are usually identified by wind tunnel tests \cite{lyu2018simulation}. 

\subsection{Differential flatness in coordinated flights}
\label{app_diff_flat}
A tail-sitter vehicle is a differentially flat system in coordinated flights, which is proven in our previous work  \cite{lu2024trajectory}. We  present some important results of the differential flatness that are used in this paper. Readers are referred to  \cite{lu2024trajectory} for thorough derivation.

Coordinated flight indicates a flight condition where an aircraft has no sideslip (e.g., $\beta = 0, \mathbf v_{a_y}^{\mathcal{B}} = 0$) \cite{clancy1975aerodynamics}. This flight condition does not restrict the tail-sitter to reach any position in the  3-D space, but significantly improves aerodynamic efficiency and moment stability during level flights that are well-studied in a fixed-wing aircraft. Coordinated flight is also beneficial for perception sensors with a small field of view (FoV), as it maintains forward visibility. Moreover, this flight condition simplifies the task of obtaining aerodynamic models, focusing only on longitudinal aerodynamic coefficients around $\beta = 0$ (see Fig. \ref{fig_aero_coef}).

The flat output is chosen as the vehicle position $\mathbf{p} \in  \mathbb{R}^3$. Utilizing $\mathbf p$ and its derivatives, all of the system states and inputs can be represented by algebra functions. In this paper, we consider the differential flatness up to the third derivative of position, $\mathbf p^{(3)}$. The system states and inputs are defined as follows:
\begin{subequations}
	\begin{align}
		\mathbf{x} &= \left(\mathbf{p} \ \ \mathbf{v} \ \ \mathbf{R}\right) \in \mathbb R^3 \times  \mathbb R^3 \times SO(3) \\
		\mathbf{u} &= \left(a_T \ \  \boldsymbol{\omega}\right) \in \mathbb R \times \mathbb R^3 
	\end{align}
\end{subequations}
and their corresponding  flatness functions are given by \cite{lu2024trajectory}:
\begin{subequations}
	\label{e_diff_flat}
	\begin{align}
		\mathbf{x} &=  \mathcal{X}(\mathbf{p}^{(0:2)})\\
		\mathbf{u} &= \mathcal{U}(\mathbf{p}^{(1:3)})
	\end{align}
\end{subequations}
where $\mathcal{X}$ and $\mathcal{U}$ denote the state and input flatness functions respectively. The calculation process of this differential flatness transform mapping is detailed in \cite{lu2024trajectory}. It is also noted from \cite{lu2024trajectory} that the mapping exhibits singularities due to the requirements of coordinated flight condition and the use of classic aerodynamic model, in which the angle of attack $\alpha$ and sideslip angle $\beta$ are invalid when airspeed is zero. Such singularities have been discussed and resolved in the previous work \cite{lu2024trajectory}, except for the following condition:
\begin{equation}
	\mathcal{S}(\mathbf x) = \Vert \dot{\mathbf{v}} - \mathbf{g} \Vert = 0
	\label{e_sing_cond}
\end{equation}
which is essentially the free-fall condition and should be actively avoided in trajectory optimization.

\bibliography{reference}
\bibliographystyle{IEEEtran}

\end{document}